\documentclass[11pt]{article}

\PassOptionsToPackage{numbers, compress}{natbib}
\usepackage[utf8]{inputenc}
\usepackage{amsfonts}
\usepackage{mathrsfs}
\usepackage{amssymb}
\usepackage{amsthm}

\usepackage{placeins}
\usepackage{libertine}
\usepackage{wrapfig}
\usepackage{graphicx}
\usepackage{multirow}
\usepackage{epstopdf}
\usepackage{multicol}
\usepackage{subfigure}
\usepackage{booktabs}

\usepackage[hyperfootnotes=false]{hyperref}
\hypersetup{
 colorlinks=false,
 citecolor=Violet,
 linkcolor=Red,
 urlcolor=Blue}
\usepackage{algorithm}%
\usepackage{algorithmic}%
\usepackage{graphicx}
\usepackage{multirow}
\usepackage{multicol}
\usepackage{caption}
\usepackage{subfigure}
\usepackage{enumitem}
\usepackage[]{natbib}
\usepackage{bbm}
\usepackage{makecell}
\usepackage{mathtools}
\usepackage[table]{xcolor}
\usepackage{url}
\usepackage{bm}
\usepackage{amsmath}
\usepackage{wrapfig}

\theoremstyle{thmstyleone}%
\newtheorem{theorem}{Theorem}
\newtheorem{proposition}[theorem]{Proposition}%

\theoremstyle{thmstyletwo}%

\theoremstyle{thmstylethree}%

\DeclareMathOperator*{\argmax}{arg\,max}

\newcommand\data[1]{{\normalfont \texttt{#1}}}

\makeatletter
\newcommand\notsotiny{\@setfontsize\notsotiny\@vipt\@viipt}
\makeatother
\DeclareMathOperator{\sgn}{sgn}
\DeclareMathOperator{\Err}{Err}

\title{Winning Prize Comes from Losing Tickets: Improve Invariant Learning by Exploring Variant Parameters for Out-of-Distribution Generalization}

\author{
  Zhuo Huang$^{1}$,  
  Muyang Li$^{1}$,   
  Li Shen$^{2}$,
  Jun Yu$^3$,\\
  Chen Gong$^4$,
  Bo Han$^5$,
  Tongliang Liu$^{1}$\\[1ex]
  \small{$^1$Sydney AI Centre, The University of Sydney;}
  \small{$^2$JD Explore Academy;} \\
  \small{$^3$University of Science and Technology of China;}
  \small{$^4$Nanjing University of Science and Technology;}\\
  \small{$^5$Hong Kong Baptist University}
\date{}
}

\begin{document}

\maketitle


%
%
%
%
%


\begin{abstract}
Out-of-Distribution (OOD) Generalization aims to learn robust models that generalize well to various environments without fitting to distribution-specific features. Recent studies based on Lottery Ticket Hypothesis (LTH) address this problem by minimizing the learning target to find some of the parameters that are critical to the task. However, in OOD problems, such solutions are suboptimal as the learning task contains severe distribution noises, which can mislead the optimization process. Therefore, apart from finding the task-related parameters (\textit{i.e.}, invariant parameters), we propose \textbf{Exploring Variant parameters for Invariant Learning (EVIL)} which also leverages the distribution knowledge to find the parameters that are sensitive to distribution shift (\textit{i.e.}, variant parameters). Once the variant parameters are left out of invariant learning, a robust subnetwork that is resistant to distribution shift can be found. Additionally, the parameters that are relatively stable across distributions can be considered invariant ones to improve invariant learning. By fully exploring both variant and invariant parameters, our EVIL can effectively identify a robust subnetwork to improve OOD generalization. In extensive experiments on integrated testbed: DomainBed, EVIL can effectively and efficiently enhance many popular methods, such as ERM, IRM, SAM, etc.
    
\end{abstract}
\newpage

\maketitle

\section{Introduction}
\begin{wrapfigure}{r}{0.48\textwidth}
    \vspace{-7mm}
  \begin{center}
    \includegraphics[width=\linewidth]{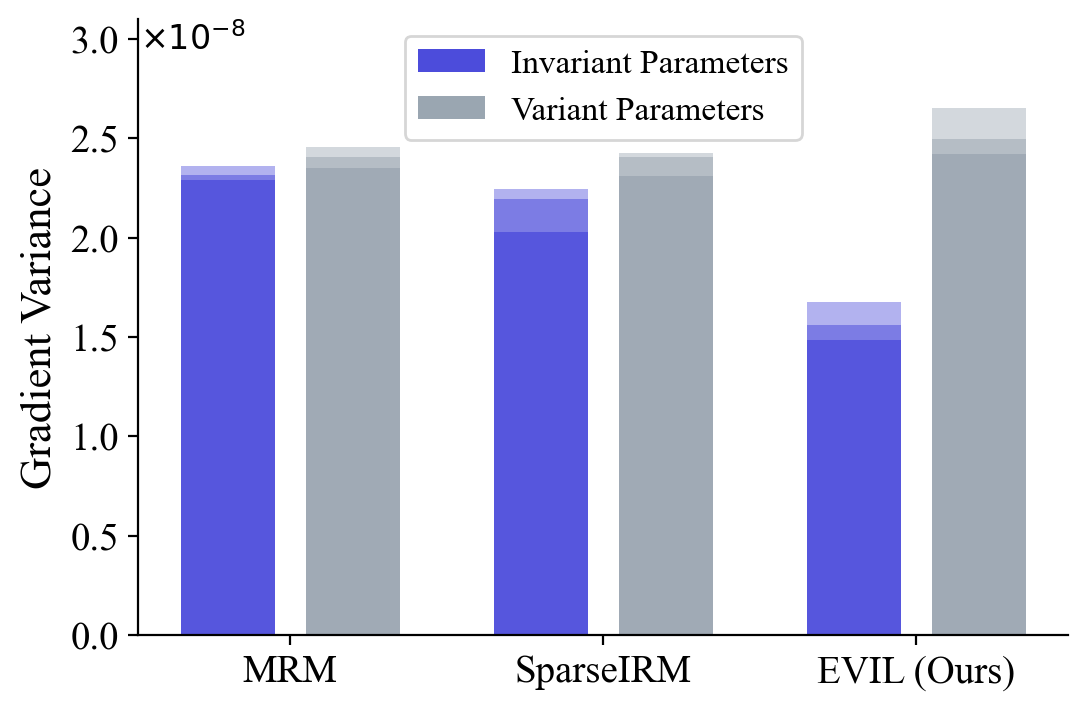}
  \end{center}
  \caption{Comparison of the gradient variance between the learned subnetwork, \textit{i.e.}, invariant parameters, and the pruned parameters, \textit{i.e.}, variant parameters. The gradient variance is computed through $\mathcal{V}=Mean(Var(\left[g_i\right]_{i=0}^d))$~\cite{rame2022fishr}, where $g_i$ denotes the $i$-th gradient among $d$ distributions, and $Var(\cdot)$ and $Mean(\cdot)$ denotes the mathematical variance and mean, respectively. The results are from three independent trials.}
\label{fig:grad_var}
\end{wrapfigure}
The strong representation ability of deep neural networks~\cite{he2016deep,krizhevsky2009learning,lecun2015deep} has been one of the vital keys to the success of deep learning over the past decade. However, the realistic deployment of neural networks is often restricted to the IID assumption where the training data and test data should be distributed independently and identically. When such an assumption is violated, a drastic degradation in learning performance is often observed which seriously hinders the practical application of deep models. Therefore, Out-Of-Distribution (OOD) generalization~\cite{gulrajani2021search, li2017deeper, muandet2013domain} thrives as a promising direction that aims to enhance model robustness against unknown distribution shifts.

In order to achieve OOD generalization, one mainstream methodology is invariant learning~\cite{arjovsky2019invariant, creager2021environment, lu2021invariant, krueger2021out} which enforces extracting invariant features to help make consistent predictions among various data distributions (or domains), meanwhile avoiding learning distribution-specific features that are irrelevant to label information. Recent advances based on Lottery Ticket Hypothesis (LTH)~\cite{frankle2019lottery, frankle2020linear, malach2020proving} shows that sparse training optimized by learning task could select some critical parameters as a subnetwork which are strongly responsible for invariant learning~\cite{morcos2019one, zhang2021can, zhou2022sparse}. However, in OOD problems, the sparsification guided by the learning task is problematic, because the distribution noise could be erroneously incorporated into the optimization of sparse learning. As a consequence, existing methods fail to identify a robust subnetwork that is stable across different distributions. Particularly, we follow Rame et al.~\cite{rame2022fishr} by using gradient variance to indicate model sensitivity to distribution shift. Then, we compare the gradient variance between the subnetwork and the pruned parameters learned by different methods, as shown in Figure~\ref{fig:grad_var}. We can see that the subnetwork learned by existing methods (MRM~\cite{zhang2021can} and SparseIRM~\cite{zhou2022sparse}) are almost as sensitive as the pruned parameters, which means that invariant information could not be fully captured.

To overcome this problem, we propose a novel sparse training framework by Exploring Variant parameters for Invariant Learning (EVIL). Specifically, by following common assumptions that input data can be decomposed into invariant features and spurious features~\cite{mitrovic2020representation, von2021self, zhang2021can, zhang2013domain}\footnote{Though some works investigate more complex situations where there are multiple factors causing the data generation process~\cite{huang2020causal, lu2021invariant, suter2019robustly}, our assumption is more common in OOD generalization.}, we can divide the network parameters into two types: \textit{invariant parameters} that are strongly related to invariant features, and \textit{variant parameters} that can mistakenly produce spurious features. Intuitively, the invariant parameters and variant parameters are mutually exclusive of each other, as they are either helpful or harmful to our learning task. In order to correctly identify an ideal subnetwork for OOD generalization, our EVIL method not only selects invariant parameters based on the learning task but also explores variant parameters via discriminating each distribution, \textit{i.e.}, classifying the data based on the distribution information. In this way, the connection between variant parameters and spurious features can be successfully established. By finding those variant parameters that are strongly activated by the distribution information, we can be sure that they should not be identified as invariant ones, which provides an alternative and effective way to improve invariant learning.

Furthermore, to dynamically improve our identification of invariant parameters during the course of network training, we propose to revisit some parameters that hardly vary when facing distribution shifts. Concretely, starting from an initialized partition of invariant parameters and variant ones, we select some variant parameters that show low response to the distribution information, as they might be critical for learning distribution-invariant features. Hence, such parameters are recollected as invariant ones to learn from label information. On the other hand, some invariant parameters that are insensitive to our learning task shall be rejected from sparse training, as they hardly contribute to invariant learning. Through this dynamic process, we are able to identify a robust subnetwork that is stable across different distributions. As shown in Figure~\ref{fig:grad_var}, the invariant parameters learned by our EVIL method show much smaller gradient variance than the rest variant parameters, which manifests its effectiveness for capturing the invariant information. 

By applying our EVIL framework to many existing OOD generalization methods, we conduct extensive empirical comparison and analysis to show that EVIL brings promising improvement with little computation cost. Specifically, when combined with simple ERM, our method achieves $2.4\%$ gains on averaged performance from DomainBed. Furthermore, our EVIL framework can surpass existing sparse training methods for invariant learning by a large margin in various sparsity levels.

To sum up, our contributions are three-fold:
\begin{itemize}
	\item[--] We propose a novel sparse training framework for OOD generalization which can fully explore the variant parameters to capture invariant information.
	
	\item[--] An iterative strategy is designed to dynamically improve the identification of robust subnetworks.
	
	\item[--] The proposed EVIL framework can be deployed to many popular methods with great effectiveness and efficiency. Moreover, EVIL effectively surpasses existing sparse invariant learning methods.
\end{itemize}

\section{Related Work}
\textbf{Invariant learning for OOD generalization} seeks to enforce model predictive invariance when facing distribution shifts~\cite{arjovsky2019invariant, huang2021universal, huang2022they, bai2022rsa}. Invariant Risk Minimization (IRM)~\cite{arjovsky2019invariant} tries to find an optimal classifier for each data distribution such that the spurious information from each domain is left out. Then, Distributionally Robust Optimization (DRO)~\cite{hu2018does, huang2023robust, sagawa2019distributionally, wang2023doe} proposes to tackle the most challenging distribution to improve OOD generalization, which is shown effective by using strong regularization penalties. Moreover, Sharpness-Aware risk Minimization (SAM)~\cite{foret2020sharpness} hopes to learn a flat loss landscape via penalizing the sharpness measurement to improve generalization results~\cite{cha2021swad, huang2023FlatMatch} and robustness to label noise~\cite{kang2023unleashing, xia2023combating, xia2021sample}. Further, Risk Extrapolation (REx)~\cite{krueger2021out} finds out that only focusing on one of the known distributions might not help generalize to unknown distributions. Instead, REx shows it is beneficial to enforce comparable performance among all training distributions via penalizing risk variance. Additionally, other methods draw insights from causality~\cite{glymour2016causal, pearl2009causality, huang2023harnessing, wang2022exploring} to disentangle the invariant features from spurious ones~\cite{gong2016domain, lu2021invariant, mitrovic2020representation} so that model prediction would not be significantly affected by distribution shift~\cite{wang2023learning, wang2022watermark}.


Nonetheless, existing invariant learning methods suffer from two major drawbacks. Firstly, some of them a computationally expensive. For example, SAM requires second-order computation to manipulate gradient information, and causality-based methods often require training generative models which is hard to be deployed on large-scale datasets. Secondly, as found out by Gulrajani et al.~\cite{gulrajani2021search}, most methods have limited performances which are even worse than Empirical Risk Minimization (ERM)! However, our EVIL can not only avoid redundant optimization on the variant parameters but also fully capture the invariant feature to achieve superior generalization accuracy.

\textbf{Sparse training for OOD generalization} is first brought out by Morcos et al.~\cite{morcos2019one}, which aims to discover the generalization ability of sparse networks obtained via common initialization methods. Then, Modular Risk Minimization (MRM)~\cite{zhang2021can} shows that sparse training can possibly improve the OOD generalization performance compared to the original dense network. However, MRM is designed in a static way which cannot be optimized along network training, hindering the sparse learning results. To tackle this issue, Sparse Invariant Risk Minimization (SparseIRM)~\cite{zhou2022sparse} proposes to conduct the sparse training process and IRM simultaneously. As a result, its generalization performance is further improved compared to MRM.

Despite the improvement of existing sparse methods, they are still suboptimal as the sparse training could be affected by the noisy gradient from the learning task. Meanwhile, the pruned parameters are not properly leveraged which would cause non-negligible information loss. Fortunately, EVIL can fully explore both variant and invariant parameters in a dynamic way. Thus, it effectively finds an ideal subnetwork that is minimally influenced by distribution shift.

\section{A Critical Analysis of Sparse Training with OOD Data}
\label{sec:theoretic_analysis}
OOD generalization aims to learn an invariant predictior by leveraging multiple distributions of training data such that the generalization performance on unseen test data distributions. Practically, we usually have multiple datasets correspondingly drawn from $m$ distributions (also termed environment), $\mathcal{E}=\{e_1, \ldots e_m\}$, where each distribution $e=\{(\mathbf{x}_i, y_i)\}_{i=0}^n$ contains $n$ examples $\mathbf{x}\in X\in \mathbb{R}$ with class label $y\in Y\in\mathbb{R}^c$. Therefore, for each example from distribution $e$, we can assign a distribution index $d\in\mathbb{R}^m$ and denote a data point as $(\mathbf{x}, y, d)$. Moreover, we have a test dataset sampled from unseen distributions $\mathcal{E}_{unseen}$ to evaluate the generalization performance of our invariant learning. Let $f_{\theta}: \mathbb{R}\rightarrow\mathbb{R}^D$ be a parameterized model with parameters $\theta\in\Theta$ which extracts feature $Z\in\mathbb{R}^M$. Our goal is to prune a sparse subnetwork from an overparameterized model so that variant features can be excluded from making the final prediction. Therefore, OOD generalization results can be improved.

\paragraph{Data Generation Process.} By following the same formulation and problem setting from Zhang et al.~\cite{zhang2021can}, we assume the input variable $X^e$ from environment $e$ is generated from latent variables $Z^e=(Z^e_{inv}, Z^e_{var})$. Intuitively, the input $X^e$ indicates the image pixels, while $Z^e_{inv}$ stands for the feature of the object-of-interest that stays invariant across different environments, and $Z^e_{var}$ denotes the spurious feature which is introduced by change of environments. Then, the data is generated through $X^e = G(Z^e_{inv}, Z^e_{var})$ where $G(\cdot)$ denotes the data generating function. To obtain an OOD-robust model, we hope to extract learning representations $Z^e$ which can recover the invariant feature $Z^e_{inv}$ meanwhile excluding variant feature $Z^e_{var}$. Such a process is modeled through $Z^e=f_{\theta}(X^e)$, where we hope $Z^e\approx \left[Z^e_{inv}, \mathbf{0}\right]$. Hence, based on the extracted feature, we can make predictions through a classification head $\hat{Y}^e=h(Z^e)$ and train our model by minimizing the error between prediction $\hat{Y}^e$ and ground truth label $Y^e$.

\paragraph{The Cause of Data Bias.} Based on the data generation process, here we explain why different distributions contain biases that hinder the generalization result. We consider a simple example where $Z^e_{inv}$ and $Z^e_{var}$ are multivariate variables with binary elements, i.e., $Z^e_{inv}\in \{-1,1\}^{M_{inv}}$ and $Z^e_{var}\in \{-1,1\}^{M_{var}}$, in which $M_{inv}$ and $M_{var}$ denotes the dimension of invariant feature and spurious feature, respectively. We have class label $Y^e\in\{-1, 1\}$ and distribution index $D\in\{-1, 1\}$. Since the invariant feature stays constant across environments, we assume each element of $Z^e_{inv}$ is equal to $Y^e$. On the other hand, we assume each element in $Z^e_{var}$ takes a value equal to $Y^e$ with probability $p^e$ and $-Y^e$ with probability $1-p^e$~\cite{zhang2021can}. When $p^e$ is large, the spurious feature would be closely correlated with the class label, hence being unlikely to introduce large data biases. Conversely, if $p^e$ is small, $Z^e_{var}$ can easily introduce noisy signals that might flip the prediction.

Additionally, we analyze the domain knowledge to provide an opposite perspective, which is overlooked by previous works~\cite{zhang2021can, zhou2022sparse}. Concretely, the change of distribution index $D$ is the cause of introducing spurious feature, i.e., $D\rightarrow Z^e_{var}$, as described by many proposed causal structures~\cite{gong2016domain, liu2021learning, sun2021recovering}. Therefore, when given the distribution $D$, we can find a specific type of spurious feature. Hence, we assume each element of $Z^e_{var}$ is equal to $D$. On the other hand, we consider $Z^e_{inv}$ takes the value of $D$ with probability $q^e$ and $-D$ with probability $1-q^e$. It has been commonly assumed that the invariant feature and domain knowledge are independent of each other~\cite{gong2016domain}, thus the probability $q^e$ could approximately be $0.5$.

\paragraph{The Flaw of Common Sparse Training Strategy}
Existing studies on sparse invariant learning~\cite{zhang2021can, zhou2022sparse} have shown that when pruning an overparameterized model, the OOD generalization performance could be improved substantially. However, we find that the existing pruning strategy that is based on ERM or objectives only related to labels could be suboptimal. Specifically, we consider the same data setting described above, $Z^e_{inv}\in\{-1, 1\}^{M_{inv}}$ and $Z^e_{var}\in\{-1, 1\}^{M_{var}}$. The data generating function $G$ is simplified as an identity map~\cite{tsipras2018robustness, rosenfeld2020risks}, thus $X=(Z^e_{inv}, Z^e_{var})$. Suppose the classification model $f_{\theta}$ is a linear layer, we have a mask $\mathbf{m}$ randomly initialized with $0-1$ values to prune the parameter $\theta$, and its sparsity ratio is set to $R=\frac{M_{var}}{M}$. Particularly, we denote the selected invariant parameters as $\theta_{inv}=\mathbf{m}\circ\theta$ and the pruned variant parameters as $\theta_{var}=(\mathbf{1}-\mathbf{m})\circ\theta$ where $\circ$ is the element-wise production. To ease the calculation, let the parameter values follow a unit norm, i.e., $\theta=\mathbf{1}\frac{1}{\sqrt{M}}$\footnote{Note that our assumption is more general than that from Zhang et al.~\cite{zhang2021can}, in which only two extreme case are considered: an optimal sparse invariant network only extracts invariant feature and a network completely depending on spurious feature. Our assumption is practical since it is similar to an initial state where all the parameters are initialized with unit-norm.}.

\begin{proposition}
	Consider a biased dataset described above, where $Z^e_{inv}\in\{-1, 1\}^{M_{inv}}$ and $Z^e_{var}\in\{-1, 1\}^{M_{var}}$, let mask $\mathbf{m}$ being randomly initialized to $0-1$ values with sparsity ratio $R=\frac{M_{var}}{M}$, and assuming $Z^e$ is a multivariate variable with independent elements. For common sparse training strategy that aims to minimize empirical risk $\Err^e=\frac{1}{2}\mathbb{E}_{(X^e, Y^e)\sim e}\left[1-Y^e\hat{Y}^e\right]$, we have:
	\begin{itemize}
		\item[--] Common strategy fails to find invariant parameters, i.e., $\mathbf{m}_{i\in\left[0, M_{inv}\right]}$ is unupdated. When leveraging domain knowledge with regularization $\Err^d=\frac{1}{2}\mathbb{E}_{(X, Y)\sim \mathcal{E}}\left[1-D\hat{D}\right]$, the invariant parameters can be effectively selected with probability at least $1-\frac{q^e}{2}$;
		\item[--] On an unknown distribution, the performance of the common strategy is highly sensitive to $p^e$: $\Err^e\le\mathcal{O}(e^{-(p^e)^4})$ while leveraging domain knowledge achieves tighter error bound when $p^e$ is small: $\Err^e\le\mathcal{O}(e^{-(p^e)^2})$.
	\end{itemize}
\end{proposition}

The detailed proof can be found in the Appendix~\ref{sec:proof}. As we find out, the pruning strategy is not sufficient to find an ideal subnetwork that can exclude spurious features meanwhile extracting invariant features. This is because the invariant parameters do not produce any error. As a result, existing strategy based on connection sensitivity~\cite{lee2018snip}, weight value~\cite{evci2020rigging}, and fisher information~\cite{sung2021training} could be suboptimal when dealing with OOD problems because the gradient information is not actually related to invariant parameters, but variant parameters. Based on this intuition, we proposed a simple yet effective strategy that leverages an additional domain knowledge regularization to explore the invariant parameters. Thanks to such a regularization, the invariant parameters can be selected because they generate gradients when calculating the distribution regularization, thus easy to find. Meanwhile, the variant parameters can still be excluded to avoid learning spurious features. Moreover, based on the error bounds, our method is insensitive to the spurious correlation $1-p^e$ compared to the common strategy. In a difficult scenario where $p^e$ is small, our method can still be robust to distribution shift.

\section{Methodology}
\begin{figure*}[t]
	\centering
	\vspace*{-8mm}
	\includegraphics[width=0.9\linewidth]{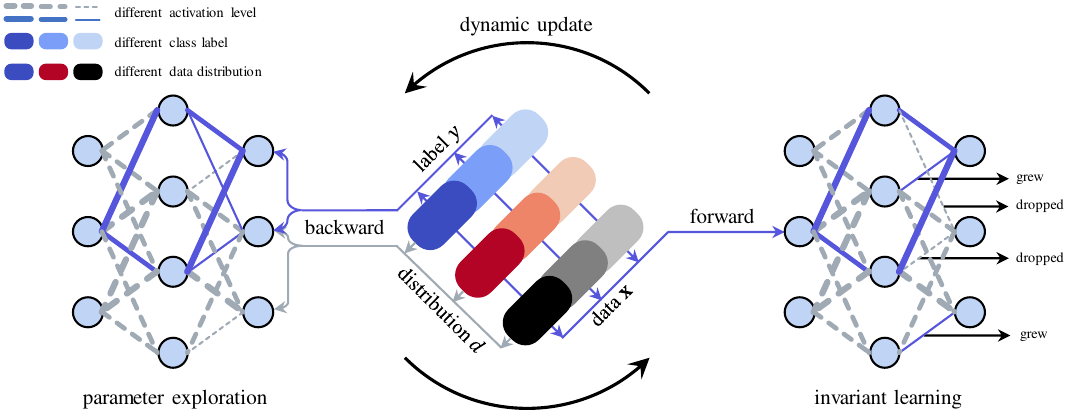}
	\caption{Learning flow of EVIL: The blocks in the middle are the training dataset where different levels of shades denote different classes, and different types of color indicate different data distributions. The blue arrows ({\color[rgb]{0.33725, 0.33725, 0.86667}$\boldsymbol{\rightarrow}$}) and gray arrows ({\color[rgb]{0.62745, 0.66667, 0.7098}$\boldsymbol{\rightarrow}$}) stand for the information flow that related to label and distribution, respectively. Moreover, the blue solid lines ({\color[rgb]{0.33725, 0.33725, 0.86667} \rule[.5ex]{1em}{1.5pt}}) and gray dashed lines ({\color[rgb]{0.62745, 0.66667, 0.7098} \rule[.5ex]{0.2em}{1.5pt}}{\color[rgb]{1, 1, 1}\rule[.5ex]{0.2em}{1.5pt}}{\color[rgb]{0.62745, 0.66667, 0.7098}\rule[.5ex]{0.2em}{1.5pt}}{\color[rgb]{1, 1, 1}\rule[.5ex]{0.2em}{1.5pt}}{\color[rgb]{0.62745, 0.66667, 0.7098}\rule[.5ex]{0.2em}{1.5pt}}) that connect neurons are the selected invariant parameters and pruned variant ones, respectively.}
	\label{fig:framework}
	\vspace*{-3mm}
\end{figure*}

In this section, we introduce our EVIL framework as shown in Figure~\ref{fig:framework}. In the learning flow of EVIL, there are two procedures: Parameter Exploration in which we propose to not only study invariant parameters but also explore the variant ones; and Invariant Learning where we train the identified subnetwork to optimize the invariant parameters. 

In the following content, we first introduce our EVIL framework which contains the aforementioned two procedures. Then, we carefully demonstrate the realization of EVIL using an important optimization method: SAM~\cite{foret2020sharpness}, which shows great improvement in OOD generalization.

\subsection{The Proposed EVIL Framework}
\label{sec:evil}
In order to get a good initialization, a few steps of pre-training are commonly conducted by minimizing a learning objective $\mathcal{L}(f_{\theta}(\mathbf{x}))$~\cite{chen2021lottery, dettmers2019sparse, evci2020rigging, morcos2019one}, where $f_{\theta}$ is a deep model with parameters $\theta\in\mathbb{R}^N$. To sparsify the deep model, a binary mask $\mathbf{m}$ is often applied through element-wise product $\mathbf{m}\circ\theta$. Such a mask $\mathbf{m}$ is either learned through optimization~\cite{csordas2020neural, dettmers2019sparse, louizoslearning, zhang2021can, zhou2022sparse}, or obtained based on certain criteria, such as connection sensitivity~\cite{lee2018snip}, weight value~\cite{evci2020rigging}, fisher information~\cite{sung2021training}, or even random initialization~\cite{frankle2019lottery, liu2022unreasonable}. By setting the sparsity ratio $R=1-\frac{\|\mathbf{m}\|_0}{|\theta|}$, we can decide how many parameters are rejected from sparse training. Then, we start from a pre-trained model with an initialized mask $\mathbf{m}$.

\paragraph{Parameter Exploration.}\label{sec:parameter_exploration} In this step, we mainly have two optimization targets:
\vspace*{-1mm}
\begin{equation}
	\min_{f_{\theta_{inv}}\otimes h} \mathcal{L}_{inv}(h(f_{\theta_{inv}}(\mathbf{x})), y),
	\label{eq:inv}
	\vspace*{-1mm}
\end{equation}
\begin{equation}
	\min_{f_{\theta_{var}}\otimes g} \mathcal{L}_{var}(g(f_{\theta_{var}}(\mathbf{x})), d),
	\label{eq:var}
\end{equation}
where $\theta_{inv}=\mathbf{m}\circ\theta$ and $\theta_{var}=(1 - \mathbf{m})\circ\theta$ denote the corresponding invariant parameters and variant ones divided by the mask $\mathbf{m}$, $h$ and $g$ are two fully-connected layers which map the extracted features into class label space $\mathbb{R}^c$ and distribution index space $\mathbb{R}^m$, respectively. Intuitively, the objective $\mathcal{L}_{inv}$ is the classification task which tries to make predictions based on the label information, and $\mathcal{L}_{var}$ tries to discriminate each distribution based on the distribution information, which is spurious and unwanted. 

By minimizing $\mathcal{L}_{inv}$ in Eq.~\ref{eq:inv}, the gradient magnitude $\nabla_{\theta_{inv}}\mathcal{L}_{inv}$~\cite{lee2018snip} can be used to find the most relevant parameters to our loss function (Note that other aforementioned sparse training criteria can be used). Similarly, by minimizing $\mathcal{L}_{var}$ in Eq.~\ref{eq:var}, those parameters with large $\nabla_{\theta_{var}}\mathcal{L}_{var}$ are sensitive to the spurious information which cannot help produce invariant features. Thus, we can sort the parameters based on the gradient magnitude to show how much they are activated by their corresponding objective.


Further, to dynamically improve our sparsification. We propose to update the mask $\mathbf{m}$ for every $\Delta T$ iterations by rejecting the least activated invariant parameters, meanwhile calling back the least activated variant parameters as invariant ones. Specifically, this process is conducted as:
\begin{equation}
	\begin{aligned}
		&\mathbf{m}\left[ArgTopK(-|\nabla_{\theta_{inv}}\mathcal{L}_{inv}|, \|\mathbf{m}\|_0S(t, \alpha, T))\right]=0,\\
		&\mathbf{m}\left[ArgTopK(-|\nabla_{\theta_{var}}\mathcal{L}_{var}|, \|\mathbf{m}\|_0S(t, \alpha, T))\right]=1,
		\label{eq:update}	
	\end{aligned}	
\end{equation}
where $ArgTopK(v, k)$ returns the indices of top-$k$ elements regarding value $v$, $\mathbf{m}\left[\cdot\right]$ denotes indexing $\mathbf{m}$. Moreover, to decide how many parameters should be exchanged, we follow Dettmers \& Zettlemoyer~\cite{dettmers2019sparse} to use cosine annealing function $S(t, \alpha, T)=\frac{\alpha}{2}(1+cos(\frac{t\pi}{T}))$, where $t$ and $T$ are the current iteration and total iterations, respectively, and hyper-parameter $\alpha$ decides the largest value. Intuitively, such a cosine annealing function gradually changes from $\alpha<1$ to 0. Through Eq.~\ref{eq:update}, the obtained new mask finds the parameters that are less affected by the distribution information and more related to our learning task than the previous one, further improving the invariant learning performance.

\paragraph{Invariant Learning.}\label{sec:invariant_learning}
\begin{wrapfigure}{r}{0.5\textwidth}
	\vspace*{-8mm}
	\begin{minipage}{0.5\textwidth}
            \renewcommand{\algorithmicrequire}{\textbf{Input:}}
            \renewcommand{\algorithmicensure}{\textbf{Output:}}
		\begin{algorithm}[H]
			\scriptsize
			\caption{EVIL}
			\label{alg:EVIL}
			\begin{algorithmic}[1]
				\REQUIRE Multiple training sets $\mathcal{E}=\{e_1, \ldots e_m\}$; Learning model $f_{\theta}$; Cosine annealing function $S(t, \alpha, T)$, mask $\mathbf{m}$ initialized based on weight value, iteration number of pre-training $T_{pre}$.
				\FOR{$t \in 0,1, \ldots,T-1$}
				\STATE Optimize via Eq.~\ref{eq:inv}; \COMMENT{\textit{\color{black!60} Invariant learning}}
				\IF {$t > T_{pre}$ and $t \% \Delta T == 0$}
				\STATE Obtain gradients of $\theta_{inv}$ and $\theta_{var}$ via Eq.~\ref{eq:inv} and Eq.~\ref{eq:var}, respectively; \COMMENT{\textit{\color{black!60} Parameter exploration}}
				\STATE Update the mask $\mathbf{m}$ via Eq.~\ref{eq:update};
				\ENDIF
				\ENDFOR
			\end{algorithmic}
		\end{algorithm}
	\end{minipage}
	\vspace*{3mm}
\end{wrapfigure}
After obtaining the updated mask $\mathbf{m}$, we then use the invariant parameters as a subnetwork to conduct invariant learning, which is generally formed as:
\vspace*{-3mm}
\begin{equation}
	\mathcal{L}_{inv}(h(f_{\theta_{inv}}(\mathbf{x})), y) = \mathcal{L}_{ce} + \lambda\mathcal{L}_{reg},
	\vspace*{-3mm}
	\label{eq:invariant_learning}
\end{equation}
where the first term is the empirical risk computed through cross-entropy loss, and the second term is the invariant learning regularization with penalty weight $\lambda$ which can be realized by many popular methods. For instance, to use IRM~\cite{arjovsky2019invariant}, we implement the regularization term as
\vspace*{-3mm}
\begin{equation}
	\mathcal{L}_{reg}=\frac{1}{mn}\sum_{\mathbf{x}\in \mathcal{E}}\|\nabla_{h|h=\mathbf{1}}\mathcal{L}_{ce}(h(f_{\theta_{inv}}(\mathbf{x})), y)\|^2.
	\vspace*{-3mm}
\end{equation}
To implement REx~\cite{krueger2021out}, we penalize the loss variance as
\vspace*{-3mm}
\begin{equation}
	\mathcal{L}_{reg}=\frac{1}{mn}\sum_{\mathbf{x}\in \mathcal{E}}Var(\{\mathcal{L}_{ce}(h(f_{\theta_{inv}}(\mathbf{x}|d)), y)\}_{d=1}^m).
	\vspace*{-3mm}
\end{equation}
Moreover, we can focus on the worst-case distribution to realize DRO~\cite{hu2018does, sagawa2019distributionally}:
\vspace*{-3mm}
\begin{equation}
	\mathcal{L}_{inv}=\min_{\theta_{inv}} \max_{e\in \mathcal{E}}\frac{1}{n}\sum_{\mathbf{x}\in e}\mathcal{L}_{ce}(h(f_{\theta_{inv}}(\mathbf{x})), y).
	\vspace*{-3mm}
\end{equation}
By combining with existing methods, their performance can be largely improved by EVIL, as shown in Section~\ref{sec:experiments}. The general process of EVIL is summarized in Algorithm~\ref{alg:EVIL}. For other detailed discussions on invariant learning methods, please refer to the \textbf{Appendix}. Next, we describe one realization of EVIL by adopting SAM optimizer~\cite{foret2020sharpness} to further improve the generalization performance.

\section{Experiment}
\label{sec:experiments}

In this section, we conduct extensive experiments to evaluate the performance of our EVIL based on a well-known testbed for OOD generalization: DomainBed~\cite{gulrajani2021search}. Specifically, we first describe the experimental setup. Then, we improve the performance of many popular invariant learning methods by deploying our EVIL framework, including ERM, IRM~\cite{arjovsky2019invariant}, REx~\cite{krueger2021out}, DRO~\cite{hu2018does, sagawa2019distributionally}, SAM~\cite{foret2020sharpness}, CORelation ALignment (CORAL)~\cite{sun2016deep}, SWAD~\cite{cha2021swad}, and MIRO~\cite{cha2022domain}. Further, we compare EVIL and its variant EVIL-SAM with other existing sparse invariant learning methods, including MRM~\cite{zhang2021can}, SparseIRM~\cite{zhou2022sparse} and report the results under different sparsity levels ($20\%$, $40\%$, $60\%$, and $80\%$). Finally, we perform various analytical experiments to validate the effectiveness and efficiency of EVIL.

\subsection{Experimental Setup}
\paragraph{Evaluation Protocol.} We follow the experimental setting of DomainBed~\cite{gulrajani2021search} to evaluation OOD generalization performance. Specifically, DomainBed contains seven benchmark datasets: \data{CMNIST}~\cite{arjovsky2019invariant} (60,000 images, 10 classes, and 3 domains), \data{RMNIST}~\cite{ghifary2015domain} (60,000 images, 10 classes, and 6 domains), \data{PACS}~\cite{li2017deeper} (9,991 images, 7 classes, 4 domains), \data{VLCS}~\cite{fang2013unbiased} (10,729 images, 5 classes, and 4 domains), \data{OfficeHome}~\cite{venkateswara2017deep} (15,588 images, 65 classes, and 4 domains), \data{Terra-\\Incognita}~\cite{beery2018recognition} (24,788 images, 10 classes, and 4 domains), \data{DomainNet}~\cite{peng2019moment} (586,575 images, 345 classes, and 6 domains), \data{WILDS}~\cite{koh2021wilds} (a testbed contains various dataset with significant distribution shfit, here we use two typical datasets: iWildCam and FMoW), \data{ImageNet}~\cite{russakovsky2015imagenet} (contain 1000 classes, here we use ImageNet dataset for fine-tuning, and use many of its variant dataset for OOD evaluation, including: \data{ImageNetV2}~\cite{recht2019imagenet}, \data{ImageNet-R}~\cite{hendrycks2021many}, \data{ImageNet-A}~\cite{hendrycks2021natural}, \data{ImageNet-Sketch}~\cite{wang2019learning}, and \data{ObjectNet}~\cite{barbu2019objectnet}). The results of \data{WILDS} and \data{ImageNet} datasets are shown in the appendix. For each benchmark dataset, we leave one domain out of the training dataset and use it as an OOD test dataset. In the main paper, we use pre-trained ResNet-50~\cite{he2016deep} as our backbone model and train them for 5,000 iterations on all datasets except \data{DomainNet}, which requires 15,000 iterations to converge. Moreover, in the appendix, we extend our method to large-scale visual recognition architecture CLIP ViT-B/16~\cite{radford2021learning}, and fine-tune the base model for OOD evaluation. The test accuracies generated by training models from the last step are provided. To avoid randomness, we conducted experiments for three independent trials.

\paragraph{Implementation Details.}
All our experiments are conducted on one single NVIDIA 3090 using PyTorch. To implement our EVIL framework, we first pre-train the models using ERM for 1,000 iterations. Then, a mask $\mathbf{m}$ is initialized based on the weight value. Specifically, by setting a sparsity ratio $R$, we can select parameters $R|\theta|$-largest weight values by setting their corresponding mask value as $1$. During parameter exploration, we first pass the gradient of invariant learning loss $\mathcal{L}_{inv}$, based on which we can sort the invariant parameters with their gradient magnitude from large to small. Then, we reject the $S(t, \alpha, T)$-least invariant parameters by setting their corresponding mask as 0. Similarly, we use the gradient of $\mathcal{L}_{var}$ to sort the variant parameters and recollect top-$S(t, \alpha, T)$ parameters. During invariant learning, we can apply the mask to parameter values as well as their corresponding gradients to conduct sparse training. Please refer to the \textbf{Appendix} for other details.

\subsection{Improving Invariant Learning Using EVIL}
\label{sec:improve_invariant_learning}
In this section, we deploy our EVIL framework to some well-known invariant learning methods and compare them with some other typical baseline methods. To conduct a fair comparison, we only considered end-to-end training on one single model, so some other methods that conduct model ensembling or averaging~\cite{cha2021swad, rame2022diverse, izmailov2018averaging, bai2021me} are not considered. Moreover, we use floating point operations per second (FLOPs) as a criterion to denote the computational efficiency by denoting the FLOPs of ERM as 1$\times$ ($7.8e10$). Practically, we set the sparsity ratio $R=60\%$, hyper-parameter $\alpha=0.2$, and $\Delta T=300$ to implement EVIL. The results are shown in Table~\ref{table:deployment}. We can see that our EVIL can effectively improve the performance of all chosen backbone methods. Particularly, on ERM, DRO, and SAM, EVIL can increase their test accuracies for $2.4\%$, $1.9\%$, and $1.9\%$, respectively. Moreover, EVIL-SAM achieves the best OOD generalization performance among all compared methods. Especially on \data{TerraIncognita} dataset, EVIL-SAM can improve the original performance of SAM for $7.2\%$, which indicates the effectiveness of EVIL in improving the performance of invariant learning. Moreover, compared to the FLOPs of all baseline methods, our EVIL shows much less computational burden, which manifests the great efficiency of our method.

\begin{table}[t]
	\setlength{\tabcolsep}{1.6mm}
	\centering
	\small
	\caption{Comparison between OOD generalization methods and our EVIL realization on some typical methods. The test accuracies on seven OOD generalization benchmarks from DomainBed are provided. We highlight the \textbf{best results} and the \underline{second best results}. The results marked with $\dagger$ denote some results are from original literature~\cite{cha2021swad}.
	}
	\label{table:deployment}
	\begin{tabular}{lccccccc|c|c}
		\toprule[1pt]
		\textbf{\notsotiny Algorithm}& \data{\notsotiny \textbf{CMNIST}} & \data{\notsotiny \textbf{RMNIST}} & \data{\notsotiny \textbf{PACS}} & \data{\notsotiny \textbf{VLCS}}& \data{\notsotiny \textbf{OfficeHome}}  & \data{\notsotiny \textbf{TerraInc}}& \data{\notsotiny \textbf{DomainNet}}& \textbf{\notsotiny Average}& \textbf{\notsotiny FLOPs} \\
		I-Mixup$^\dagger$~\cite{xu2020adversarial}  & 31.3  & 97.8   & 84.6 & 77.4  & 68.1& 47.9& 39.2 & 63.7     & -      \\ 
		MLDG$^\dagger$~\cite{li2018learning}  &  36.9  & 98.0   & 84.9  & 77.2  & 66.8& 47.8& 41.2 & 64.6 & -\\
		MMD$^\dagger$~\cite{li2018domain}  & \textbf{42.6}  & 98.1  &   84.7  & 77.5  & 66.4& 42.2& 23.4 & 62.1 & -\\
		DANN$^\dagger$~\cite{ganin2015unsupervised} & 29.0  &  89.1  &  83.7  & 78.6  & 65.9& 46.7& 38.3 & 61.6 & -\\
		CDANN$^\dagger$~\cite{long2018conditional}  & 31.1  &  96.3   &  82.6  & 77.5  & 65.7& 45.8& 38.3 & 62.5 & -\\
		MTL$^\dagger$~\cite{blanchard2021domain}   &  30.4  & 97.2   & 84.6  & 77.2  & 66.4& 45.6& 40.6 & 63.1 & -\\
		SagNet$^\dagger$~\cite{nam2021reducing}  &  34.2   & 96.4  & 86.3  & 77.8  & 68.1& 48.6& 40.3 & 64.5 & -\\
		ARM$^\dagger$~\cite{zhang2020adaptive}   &  32.6   & 98.1   & 85.1  & 77.6  & 64.8& 45.5& 35.5 & 62.5 & -\\
		RSC$^\dagger$~\cite{huang2020self}   & 35.2  & 96.3   & 85.2  & 77.1  & 65.5& 46.6& 38.9 & 63.5 & -\\
		Mixstyle$^\dagger$~\cite{zhou2021domain} &  38.5   &  97.2   & 85.2  & 77.9  & 60.4& 44.0& 34.0 & 62.4 & -\\  \midrule
		ERM$^\dagger$~\cite{vapnik1999overview}  & 34.2  & 98.0  & 83.3  & 76.8  & 67.3& 46.2& 40.8 & 63.8 & 1$\times$\\ 
		EVIL    &  39.4   & 98.4   &  86.0& 78.8 & 68.2 & 49.1& 43.8     & 66.2 & 0.42$\times$\\  \addlinespace[-0.1cm]
		& \notsotiny{($\pm1.1$)} & \notsotiny{($\pm0.1$)} & \notsotiny{($\pm0.1$)} & \notsotiny{($\pm0.2$)} & \notsotiny{($\pm0.2$)} & \notsotiny{($\pm0.2$)} & \notsotiny{($\pm0.3$)} & {\color[rgb]{0.33725, 0.33725, 0.86667}\notsotiny{($\bf\uparrow2.4$)}} & \\ \midrule
		DRO$^\dagger$~\cite{sagawa2019distributionally} &  32.2  &   97.9  & 84.4  & 76.7  & 66.0& 43.2& 33.3 & 61.9 & 1$\times$\\
		EVIL-DRO   &  34.2  & 98.2   & 85.6 & 77.7  & 66.4 & 49.1     & 35.5     & 63.8 & 0.42$\times$\\  \addlinespace[-0.1cm]
		& \notsotiny{($\pm1.7$)} & \notsotiny{($\pm0.1$)} & \notsotiny{($\pm0.2$)} & \notsotiny{($\pm0.2$)} & \notsotiny{($\pm0.1$)} & \notsotiny{($\pm0.2$)} & \notsotiny{($\pm0.2$)} & {\color[rgb]{0.33725, 0.33725, 0.86667}\notsotiny{($\bf\uparrow1.9$)}} & \\ \midrule
		IRM$^\dagger$~\cite{arjovsky2019invariant}   &  36.3 &  97.7   & 83.5  & 78.6  & 64.3& 47.6& 33.9 & 63.1 & 1$\times$      \\
		EVIL-IRM  & 39.1  &  98.3    & 85.1 & 78.8  & 66.4 & 48.3     & 36.0& 64.6 &  0.42$\times$       \\ \addlinespace[-0.1cm]
		& \notsotiny{($\pm2.2$)} & \notsotiny{($\pm0.2$)} & \notsotiny{($\pm0.1$)} & \notsotiny{($\pm0.1$)} & \notsotiny{($\pm0.1$)} & \notsotiny{($\pm0.3$)} & \notsotiny{($\pm0.3$)} & {\color[rgb]{0.33725, 0.33725, 0.86667}\notsotiny{($\bf\uparrow1.5$)}} & \\ \midrule
		REx$^\dagger$~\cite{krueger2021out}  &  39.2  &  97.3   & 84.9  & 78.3  & 66.4& 46.4& 33.6 & 63.7 & 1$\times$\\
		EVIL-REx  & 41.2  &  \underline{98.7}   & 86.0 & 79.1 & 68.0& 48.4     & 34.5& 65.1    & 0.42$\times$\\  \addlinespace[-0.1cm]
		& \notsotiny{($\pm1.3$)} & \notsotiny{($\pm0.1$)} & \notsotiny{($\pm0.1$)} & \notsotiny{($\pm0.2$)} & \notsotiny{($\pm0.2$)} & \notsotiny{($\pm0.3$)} & \notsotiny{($\pm0.1$)} & {\color[rgb]{0.33725, 0.33725, 0.86667}\notsotiny{($\bf\uparrow1.4$)}} & \\ \midrule
		CORAL$^\dagger$~\cite{sun2016deep} &   29.9  &  98.1  &  86.2  & 78.8  & 68.7& 47.7& 41.5 & 64.4 &   1$\times$      \\ 
		EVIL-CORAL &  34.5 &   98.6    & 86.9 & \underline{79.2} & \underline{69.0}& 49.2     & 42.6 & 65.7 & 0.43$\times$\\  \addlinespace[-0.1cm]
		& \notsotiny{($\pm1.9$)} & \notsotiny{($\pm0.1$)} & \notsotiny{($\pm0.2$)} & \notsotiny{($\pm0.1$)} & \notsotiny{($\pm0.1$)} & \notsotiny{($\pm0.2$)} & \notsotiny{($\pm0.3$)} & {\color[rgb]{0.33725, 0.33725, 0.86667}\notsotiny{($\bf\uparrow1.3$)}} & \\ \midrule
		SWAD~\cite{cha2021swad}  &  38.3  &   98.1  &  88.1& 79.1& 70.6& 50.0& 46.5& 67.2 & 1$\times$    \\
		EVIL-SWAD & 38.7&  98.3 & \textbf{88.3} & 79.3 & \textbf{71.7} &\textbf{51.2} & \textbf{46.9} & \textbf{67.7} & 0.43$\times$\\  \addlinespace[-0.1cm]
		& \notsotiny{($\pm2.3$)} & \notsotiny{($\pm0.3$)} & \notsotiny{($\pm0.1$)} & \notsotiny{($\pm0.1$)} & \notsotiny{($\pm0.2$)} & \notsotiny{($\pm0.3$)} & \notsotiny{($\pm0.2$)} & {\color[rgb]{0.33725, 0.33725, 0.86667}\notsotiny{($\bf\uparrow0.5$)}} & \\ \midrule
		MIRO~\cite{cha2022domain}  &  39.4  &   97.5   &  85.4&  79.0&  70.5&  50.4&  44.3&  66.6 & 1$\times$    \\
		EVIL-MIRO  &  40.2  &   98.6  &  85.8&  79.4&  \underline{71.2}&  \underline{50.9}&   \underline{45.0}&  67.3 & 0.45$\times$\\  \addlinespace[-0.1cm]
		& \notsotiny{($\pm2.3$)} & \notsotiny{($\pm0.3$)} & \notsotiny{($\pm0.1$)} & \notsotiny{($\pm0.1$)} & \notsotiny{($\pm0.2$)} & \notsotiny{($\pm0.3$)} & \notsotiny{($\pm0.2$)} & {\color[rgb]{0.33725, 0.33725, 0.86667}\notsotiny{($\bf\uparrow0.7$)}} & \\ \midrule
		SAM~\cite{foret2020sharpness}  &  38.5  &   98.1    & 85.8  & 79.4 & 69.6& 43.3& 44.3 & 65.6 & 2$\times$    \\
		EVIL-SAM  & \underline{40.4}  &   \textbf{98.8}   & \underline{87.8} & \textbf{80.1} & 70.3 &  50.5   &  \underline{45.0}& \underline{67.5} & 0.89$\times$\\  \addlinespace[-0.1cm]
		& \notsotiny{($\pm2.3$)} & \notsotiny{($\pm0.3$)} & \notsotiny{($\pm0.1$)} & \notsotiny{($\pm0.1$)} & \notsotiny{($\pm0.2$)} & \notsotiny{($\pm0.3$)} & \notsotiny{($\pm0.2$)} & {\color[rgb]{0.33725, 0.33725, 0.86667}\notsotiny{($\bf\uparrow1.9$)}} & \\ 
		
	\toprule[1pt]
	\end{tabular}
	\vspace*{-2mm}
\end{table}
\FloatBarrier

\subsection{Comparing EVIL to Sparse Invariant Learning}
\label{sec:comparing_evil_to_sparse}
Furthermore, to show that our method finds a more robust subnetwork for OOD generalization, we compare EVIL with two existing sparse invariant learning methods. Specifically, to validate the effectiveness of our method under different levels of sparsity, we vary to sparsity ratio as $20\%$, $40\%$, $60\%$, and $80\%$. The experimental results are shown in Table~\ref{table:sparse_comparison}. Generally, we can see that our EVIL and EVIL-SAM surpass both MRM and SparseIRM in almost all scenarios. Moreover, among all sparsity levels, both EVIL and the other two methods achieve the best results under sparsity $60\%$. Specifically, EVIL-SAM shows the best performance under sparsity $60\%$ on almost all datasets, and it surpasses the second-best opponent for $2.8\%$ on averaged accuracy. Besides, EVIL implementation with just ERM can also improve the second-best methods for $1.5\%$ on the averaged results. As for the computational efficiency, our EVIL is comparable to other sparse training methods, except EVIL-SAM which requires an extra backward to compute the parameter perturbation. Therefore, by exploring the variant parameters, our method can successfully achieve superior OOD generalization performance with comparable efficiency to the sparse invariant learning methods.

\begin{table*}[h]
	\vspace*{2mm}
	\setlength{\tabcolsep}{1.45mm}
	\centering
	\small
	\caption{Comparison between existing sparse invariant learning methods and EVIL varying sparsity levels. The test accuracies on seven OOD generalization benchmarks from DomainBed are provided. We highlight the \textbf{best results} and the \underline{second best results}. 
	}
	\label{table:sparse_comparison}
	\begin{tabular}{clccccccc|c|c}
		\toprule[1pt]
		\textbf{\notsotiny $R$} & \textbf{\notsotiny Algorithm}& \data{\notsotiny \textbf{CMNIST}} & \data{\notsotiny \textbf{RMNIST}} & \data{\notsotiny \textbf{PACS}} & \data{\notsotiny \textbf{VLCS}}& \data{\notsotiny \textbf{OfficeHome}}  & \data{\notsotiny \textbf{TerraInc}}& \data{\notsotiny \textbf{DomainNet}}& \textbf{\notsotiny Average}& \textbf{\notsotiny FLOPs} \\
		\multirow{4}{*}{20\%} & {\notsotiny MRM}~\cite{zhang2021can}&  31.5   & 89.3  & 78.3& 70.0& 63.6 & 42.7 & 37.9& 59.0& 0.82$\times$ \\ 
		& {\notsotiny SparseIRM}~\cite{zhou2022sparse}&  31.5   & 92.2 & 80.8& 71.2& 63.7 & 43.0 & 39.6& 60.3 & 0.81$\times$\\
		& {\notsotiny EVIL}&  34.1   & 95.8  & 82.2& 73.7& 66.3 & 45.3 & 41.9& 62.7 & 0.82$\times$ \\ \addlinespace[-0.1cm]
		& & \notsotiny{($\pm1.7$)} & \notsotiny{($\pm0.1$)}& \notsotiny{($\pm0.2$)} & \notsotiny{($\pm0.2$)} & \notsotiny{($\pm0.2$)} & \notsotiny{($\pm0.3$)} & \notsotiny{($\pm0.2$)} &  & \\
		& {\notsotiny EVIL-SAM}&  35.2   & 96.3  & 83.6& 74.0& 65.6 & 45.3 & 42.6& 63.2 &  1.62$\times$\\ \addlinespace[-0.1cm]
		& & \notsotiny{($\pm2.3$)} & \notsotiny{($\pm0.2$)}& \notsotiny{($\pm0.2$)} & \notsotiny{($\pm0.2$)} & \notsotiny{($\pm0.2$)} & \notsotiny{($\pm0.1$)} & \notsotiny{($\pm0.1$)} &  & \\ \midrule
		\multirow{4}{*}{40\%} & {\notsotiny MRM}~\cite{zhang2021can} &  36.2   & 95.8 & 81.9& 73.5& 63.1 & 45.6 & 40.4& 62.3& 0.62$\times$ \\ 
		& {\notsotiny SparseIRM}~\cite{zhou2022sparse}&  35.7   & 96.4 & 82.5& 74.2& 66.8 & 47.8 & 42.6& 63.7 & 0.62$\times$\\
		& {\notsotiny EVIL} &  38.9   & 97.3 & 84.7& 75.3& 66.4 & 47.1 & 44.0& 64.8 & 0.62$\times$\\ \addlinespace[-0.1cm]
		& & \notsotiny{($\pm1.6$)} & \notsotiny{($\pm0.2$)}& \notsotiny{($\pm0.1$)} & \notsotiny{($\pm0.2$)} & \notsotiny{($\pm0.1$)} & \notsotiny{($\pm0.1$)} & \notsotiny{($\pm0.1$)} &  & \\
		& {\notsotiny EVIL-SAM}&  38.8   & 97.9 & 84.8& 77.4& 66.9 & 48.1 & \textbf{45.2}&  65.6 & 1.33$\times$\\  \addlinespace[-0.1cm]
		& & \notsotiny{($\pm2.4$)} & \notsotiny{($\pm0.3$)}& \notsotiny{($\pm0.3$)} & \notsotiny{($\pm0.2$)} & \notsotiny{($\pm0.1$)} & \notsotiny{($\pm0.2$)} & \notsotiny{($\pm0.2$)} &  & \\ \midrule
		\multirow{4}{*}{60\%} & {\notsotiny MRM}~\cite{zhang2021can}&  38.2   & 97.6  & 83.6& 76.8& 66.5 & 46.7 & 40.3& 64.2& 0.41$\times$ \\ 
		& {\notsotiny SparseIRM}~\cite{zhou2022sparse}&  37.9   & 97.9 & 84.9& 77.3& 65.1 & 48.8 & 42.0& 64.8 & 0.42$\times$\\
		& {\notsotiny EVIL} &  \underline{39.4}   & \underline{98.4}   &  \underline{86.0}& \underline{78.8} & \underline{68.2} & \underline{49.1} & 43.8     & \underline{66.3} & 0.42$\times$\\  \addlinespace[-0.1cm]
		& & \notsotiny{($\pm1.4$)} & \notsotiny{($\pm0.2$)}& \notsotiny{($\pm0.1$)} & \notsotiny{($\pm0.2$)} & \notsotiny{($\pm0.2$)} & \notsotiny{($\pm0.2$)} & \notsotiny{($\pm0.3$)} &  & \\
		& {\notsotiny EVIL-SAM} & \textbf{40.4}  &   \textbf{98.8}   & \textbf{87.8} & \textbf{80.1} & \textbf{70.3}& \textbf{50.5}   & \underline{45.0}& \textbf{67.6} & 0.89$\times$\\  \addlinespace[-0.1cm]
		& & \notsotiny{($\pm2.2$)} & \notsotiny{($\pm0.1$)}& \notsotiny{($\pm0.1$)} & \notsotiny{($\pm0.1$)} & \notsotiny{($\pm0.2$)} & \notsotiny{($\pm0.1$)} & \notsotiny{($\pm0.2$)} &  & \\  \midrule
		\multirow{4}{*}{80\%} & {\notsotiny MRM}~\cite{zhang2021can}&  37.7   & 96.3  & 80.3& 72.0& 61.2 & 42.7 & 35.4& 60.8& 0.21$\times$ \\ 
		& {\notsotiny SparseIRM}~\cite{zhou2022sparse}&  37.8   & 97.2 & 82.9& 71.6& 62.4 & 43.8 & 36.2& 61.7 & 0.21$\times$\\
		& {\notsotiny EVIL}&  38.5   & 98.1  & 84.7& 74.1& 64.3 & 46.4 & 40.1& 63.7 & 0.21$\times$\\ \addlinespace[-0.1cm]
		& & \notsotiny{($\pm1.3$)} & \notsotiny{($\pm0.2$)}& \notsotiny{($\pm0.0$)} & \notsotiny{($\pm0.2$)} & \notsotiny{($\pm0.2$)} & \notsotiny{($\pm0.1$)} & \notsotiny{($\pm0.0$)} &  & \\
		& {\notsotiny EVIL-SAM}&  38.9  & 98.3 & \textbf{87.8}& 76.8& 65.7 & 47.6 & 42.7& 65.4 & 0.57$\times$\\ \addlinespace[-0.1cm]
		& & \notsotiny{($\pm1.2$)} & \notsotiny{($\pm0.3$)} & \notsotiny{($\pm0.2$)} & \notsotiny{($\pm0.2$)} & \notsotiny{($\pm0.2$)} & \notsotiny{($\pm0.1$)} & \notsotiny{($\pm0.3$)} &  & \\
	
		\toprule[1pt]
	\end{tabular}
\end{table*}
\FloatBarrier

\begin{figure*}[t]
	\vspace*{-5mm}
	\centering
	\includegraphics[width=\linewidth]{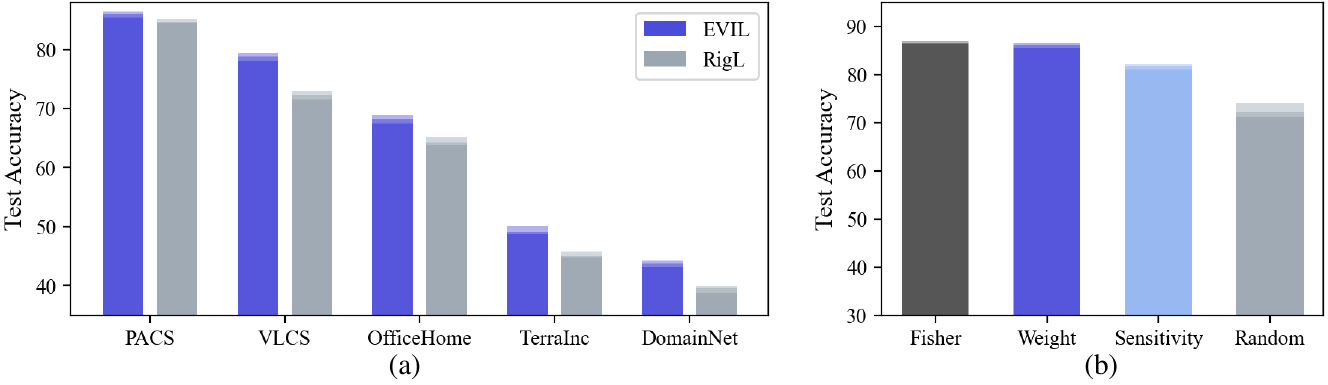}
	\vspace{-4mm}
	\caption{(a) Comparison of EVIL and RigL which leverages the label information to explore the variant parameters. (b) Comparison of different mask initialization strategies.}
	\label{fig:ablation}
\end{figure*}
\vspace{-3mm}

\subsection{Performance Analysis}
\label{sec:performance_analysis}
In this section, we conduct extensive empirical analyses to exploit why EVIL can achieve effective results. First, we conduct ablation studies to show the effect of leveraging distribution knowledge and the influence of choosing different mask initialization strategies. Then, we conduct parameter sensitivity analysis on the hyper-parameter $\alpha$ and $\Delta T$. Further, we show compare EVIL-SAM and SAM by visualizing their sharpness during training. Finally, we show the Hessian spectra to explain why EVIL can achieve good generalization results.

\paragraph{Ablation Study.} To validate the effectiveness of exploring variant parameters using distribution knowledge, we change the optimization target $\mathcal{L}_{var}(g(f_{\theta_{var}}(\mathbf{x})), d)$ in Eq.~\ref{eq:var} to $\mathcal{L}_{ce}(h(f_{\theta_{var}}(\mathbf{x})), y)$ which leverages the label information instead. As a result, the changed variant is actually Rigging the Lottery (RigL)~\cite{evci2020rigging} which is an effective sparse training method. By comparing EVIL and RigL on DomainBed as shown in Figure~\ref{fig:ablation} (a), we can see that EVIL surpasses RigL in all scenarios. Therefore, we can conclude that exploring variant parameters by using the distribution information is essential for OOD generalization. Moreover, to show the influence of choosing different mask initialization strategies as mentioned in Section~\ref{sec:evil}, we compare the weight value (as done in our method) with Fisher information~\cite{sung2021training}, connection sensitivity~\cite{lee2018snip}, and random initialization~\cite{frankle2019lottery} and show the result in Figure~\ref{fig:ablation} (b). As we can see, the Fisher information and weight value are two better strategies than connection sensitivity and random initialization, which supports our choice of using weight value.

\begin{figure}[h]
	\vspace*{-5mm}
	\begin{minipage}{0.55\linewidth}
		\includegraphics[width=\textwidth]{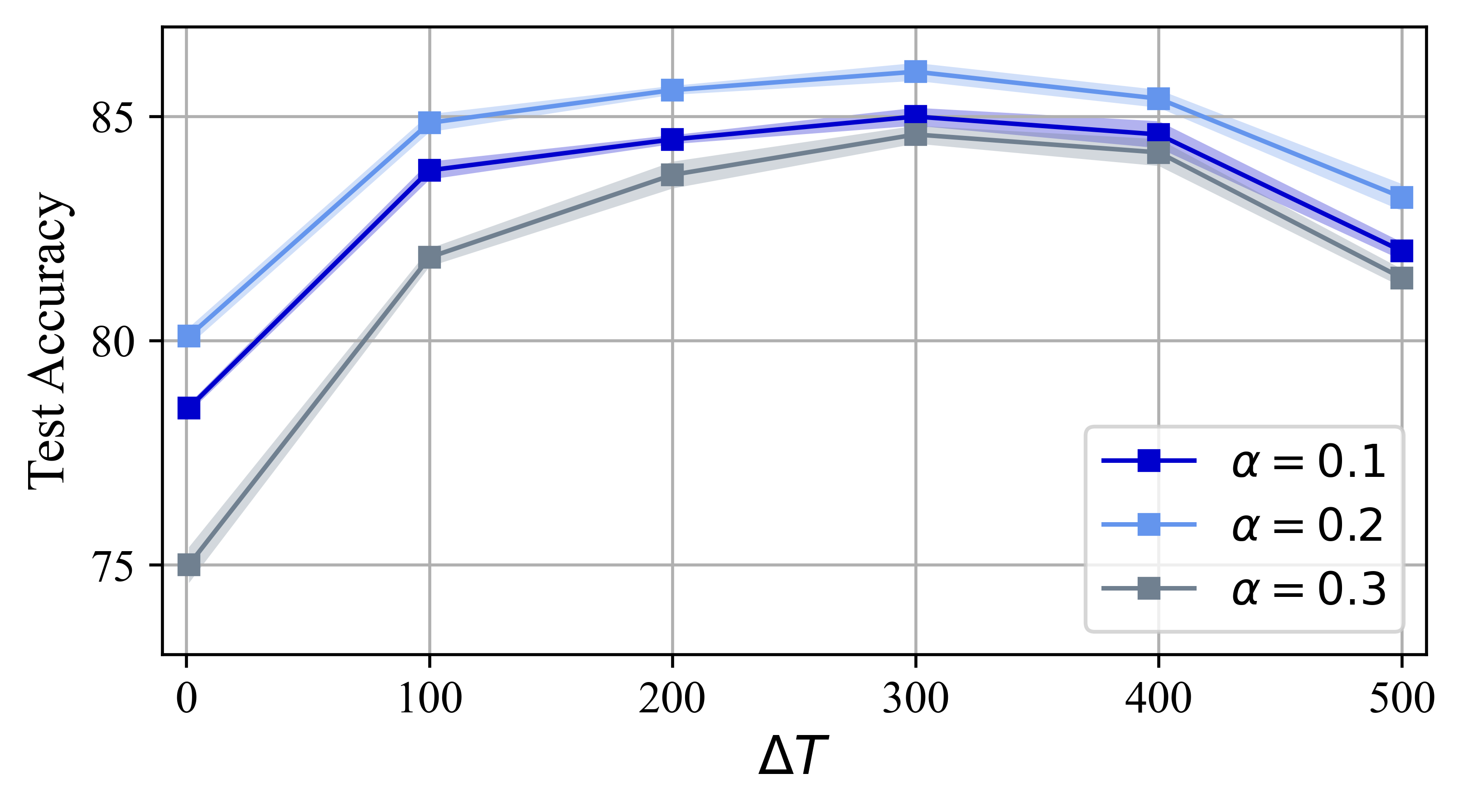}
		\vspace{-5mm}
		\caption{Parameter sensitivity analysis on $\alpha$ and $\Delta T$.}
		\label{fig:parameter}
	\end{minipage}
	\begin{minipage}{0.45\linewidth}
		\includegraphics[width=\textwidth]{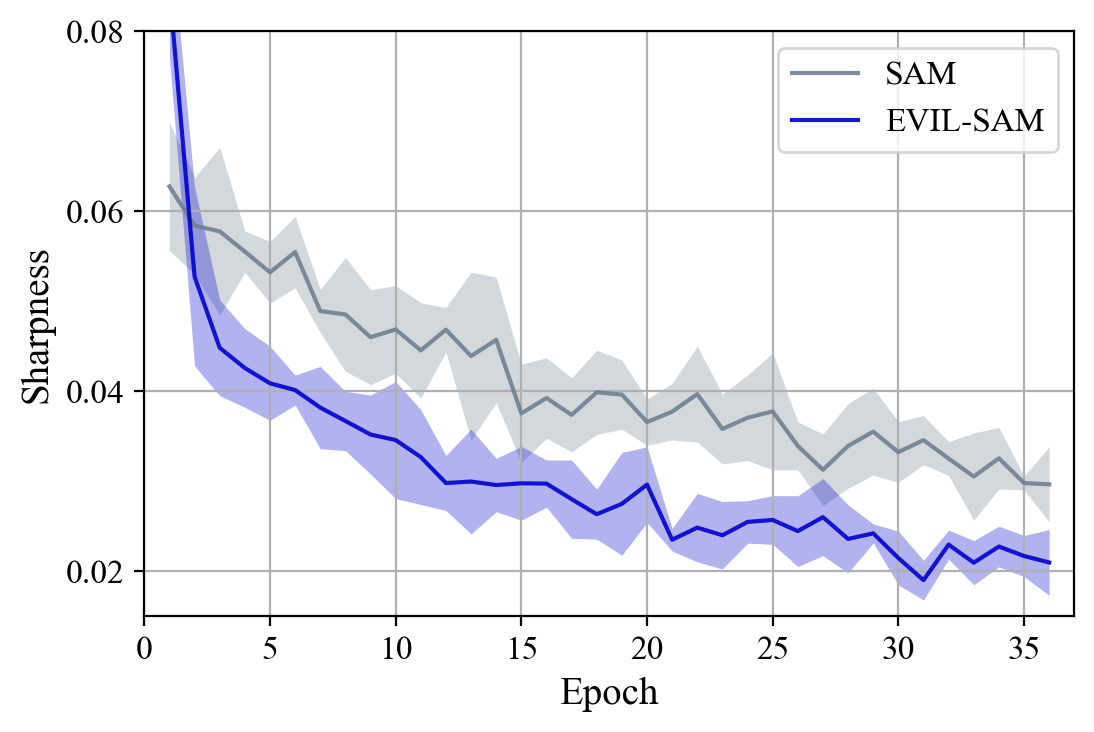}
		\vspace{-5mm}
		\caption{Sharpness comparison of EVIL-SAM and SAM.}
		\label{fig:sharpness}
	\end{minipage}
	\vspace{-5mm}
\end{figure}

\paragraph{Parameter Sensitivity Analysis.}
To analyze the different choices of $\alpha$ and $\Delta T$, we set $\alpha$ to 0.1, 0.2, and 0.3 and vary $\Delta T$ as 1, 100, 200, 300, 400, and 500. As shown in Figure~\ref{fig:parameter}, choosing $\alpha$ as 0.2, and $\Delta T$  as 300 is the best. Moreover, lower $\alpha$ would hinder the dynamic update of the mask, and high $\alpha$ would cause incorporating noises, thus both choices lead to a performance drop. On the other hand, $\Delta T$ controls the updating frequency. Both too small $\Delta T$ and too large $\Delta T$ would correspondingly cause insufficient update and redundant update, further leading to degradation.

\begin{wrapfigure}{r}{0.5\textwidth}
	\vspace{-5mm}
	\centering
	\includegraphics[width=\linewidth]{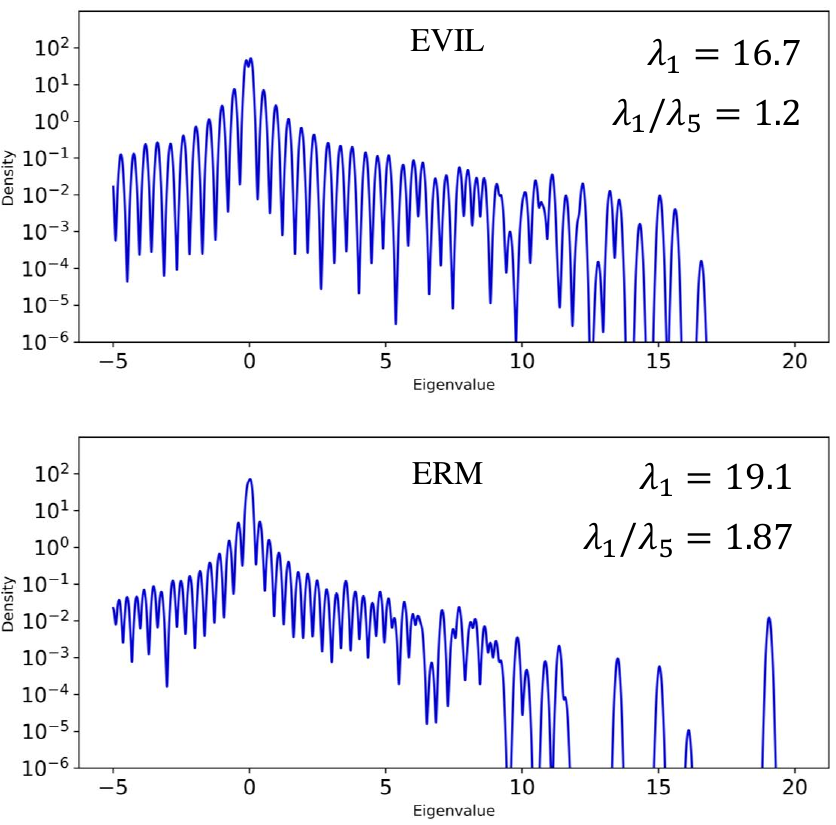}
	\vspace{-6mm}
	\caption{Hessian Spectrum of EVIL and ERM.}
	\label{fig:hessian}
	\vspace{-6mm}
\end{wrapfigure}
\paragraph{Sharpness Analysis.} To show how our EVIL affects SAM during OOD generalization, we visualize the sharpness obtained from the training process in Figure~\ref{fig:sharpness}. As a result, our EVIL-SAM can produce smaller sharpness during training than SAM, which indicates that EVIL-SAM is more robust than SAM in OOD generalization problems.

\paragraph{Hessian Spectra.}
To analyze whether an algorithm can converge to a flat minima, the Hessian spectrum is commonly used as a criterion~\cite{ghorbani2019investigation}. Specifically, we follow Foret et al.~\cite{foret2020sharpness} to use the ratio of dominant eigenvalue to fifth largest eigenvalue, \textit{i.e.}, $\lambda_1 / \lambda_5$ as the criterion for comparing EVIL and ERM. Generally, a smaller $\lambda_1 / \lambda_5$ often means a flatter minima is found. Thus, we follow Ghorbani et al.~\cite{ghorbani2019investigation} by using the Lanczos algorithm to approximate the Hessian spectrums of ERM and EVIL in Figure~\ref{fig:hessian}. As we can see the $\lambda_1 / \lambda_5$ of EVIL is much smaller than that of ERM, which confirms that our method can converge to a flatter minima than ERM. Moreover, as the dominant eigenvalue $\lambda_1$ is also an important measurement, we can see that EVIL produces smaller $\lambda_1$ than ERM as well, which again supports the effectiveness of EVIL. Therefore, it is reasonable that EVIL can achieve great generalization results.

\section{Conclusion}
In this paper, we aim to address the problem that existing sparse invariant learning methods fail to fully capture invariant information in OOD generalization problems, owing to the misleading influence of distribution shifts. Therefore, we propose EVIL by leveraging the distribution knowledge to explore the variant parameters. By finding the variant parameters that are highly sensitive to distribution shift, we can identify a robust subnetwork that effectively extracts invariant features. Moreover, we propose to improve our identification dynamically during network training. As a result, our EVIL framework can effectively and efficiently improve the OOD generalization performance of many invariant learning methods, meanwhile surpassing all compared sparse invariant learning methods. Exhaustive analyses are conducted to comprehensively validate the performance of EVIL.

\newpage
\appendix
\label{sec:appendix}

\begin{center}
	{\Large\bf Appendix}
\end{center}
\vspace{0.3in}

In this Appendix, we first provide detailed proof of our proposition in Section~\ref{sec:theoretic_analysis}. Then, we provide extra implementation details. Finally, we conduct additional experiments regarding extra invariant learning methods, detailed analysis on SAM optimizer~\cite{fort2021exploring}, various network architectures such as the most sophisticated visual recognition model CLIP ViT-B/16~\cite{radford2021learning}, and large-scale datasets including WILDS~\cite{koh2021wilds} and ImageNet~\cite{russakovsky2015imagenet}.

\section{Proof of Proposition 1}
\label{sec:proof}
First, by applying the selected invariant parameters, we can have the label prediction $\hat{Y}^e=\sgn(\theta_{inv}^\top Z^e)=\sgn((\mathbf{m}\circ\theta)^\top Z^e)$ where $\sgn(\cdot)$ returns the sign of input value. Further, we have:
\vspace*{-5mm}
\begin{align}
	(\mathbf{m}\circ\theta)^\top Z^e &=\frac{1}{\sqrt{M}}\mathbf{m}^\top Z^e\\
	\notag
	&= \sqrt{M}\frac{1}{M}\left[\mathbf{m}_{i\in\left[0, M_{inv}\right]}, \mathbf{m}_{i\in\left[M_{inv}, M\right]}\right]^\top \left[Z^e_{inv}, Z^e_{var}\right]\\
	&= \sqrt{M}\left[\frac{1}{M_{inv}}\mathbf{m}_{i\in\left[0, M_{inv}\right]}^\top Z^e_{inv}+\frac{1}{M_{var}}\mathbf{m}_{i\in\left[M_{inv}, M\right]}^\top Z^e_{var}\right].\label{eq:appendix_matrix_multiply}
\end{align}
Then the error produced by the current sparse network for a given environment is:
\begin{align}
	\Err^e&=\frac{1}{2}\mathbb{E}_{(X^e, Y^e)\sim e}\left[1-Y^e\hat{Y}^e\right]=\frac{1}{2}\left[1-\mathbb{E}^e\left[Y^e\hat{Y}^e\right]\right].
\end{align}
Here we simplify the expectation on samples from distribution $e$ as $\mathbb{E}^e$.
\begin{align}
	\notag
	\mathbb{E}^e\left[Y^e\hat{Y}^e\right]&=\mathbb{E}^e\left[\sgn(\frac{1}{M}\mathbf{m}^\top Z^e)Y^e\right]\\
	&=\sum_{y\in\left\{-1, 1\right\}}\mathbb{P}\left[Y^e=y\right]\mathbb{E}^e\left[\sgn(\frac{1}{M}\mathbf{m}^\top Z^e)\mid Y^e=y\right]y\label{eq:appendix_expected_prediction}.
\end{align}
\begin{align}
	\notag
	\mathbb{E}^e\left[\sgn(\frac{1}{M}\mathbf{m}^\top Z^e)\mid Y^e=1\right]&=\mathbb{P}\left[\frac{1}{M}\mathbf{m}^\top Z^e>0\mid Y^e=1\right]-\mathbb{P}\left[\frac{1}{M}\mathbf{m}^\top Z^e\le0\mid Y^e=1\right]\\
	&=2\mathbb{P}\left[\frac{1}{M}\mathbf{m}^\top Z^e>0\mid Y^e=1\right]-1\label{eq:appendix_expected_prediction_one}.
\end{align}
Similar to Zhang et al.~\cite{zhang2021can}, we observe that $\mathbb{P}\left[\frac{1}{M}\mathbf{m}^\top Z^e\le0\mid Y^e=1\right]=\mathbb{P}\left[\frac{1}{M}\mathbf{m}^\top Z^e>0\mid Y^e=-1\right]$, and plug Eq.~\ref{eq:appendix_expected_prediction_one} into Eq.~\ref{eq:appendix_expected_prediction} to get:
\begin{align}
	\Err^e=\mathbb{P}\left[\frac{1}{M}\mathbf{m}^\top Z^e\le0\mid Y^e=1\right]\label{eq:appendix_expected_error}.
\end{align}
Similar to Eq.~\ref{eq:appendix_matrix_multiply}, we further decompose Eq.~\ref{eq:appendix_expected_error}:
\begin{align}
	\notag
	\Err^e&=\mathbb{P}\left[\frac{1}{M_{inv}}\mathbf{m}_{i\in\left[0, M_{inv}\right]}^\top Z^e_{inv}+\frac{1}{M_{var}}\mathbf{m}_{i\in\left[M_{inv}, M\right]}^\top Z^e_{var}\le0\mid Y^e=1\right]\\
	&\le\mathbb{P}\left[\frac{1}{M_{inv}}\mathbf{m}_{i\in\left[0, M_{inv}\right]}^\top Z^e_{inv}\le0\mid Y^e=1\right]+\mathbb{P}\left[\frac{1}{M_{var}}\mathbf{m}_{i\in\left[M_{inv}, M\right]}^\top Z^e_{var}\le0\mid Y^e=1\right].
\end{align}
Since all elements in $Z^e_{inv}$ are equal to $Y^e$, the first term equals $0$, thus the equality also holds. Then, we have:
\begin{align}
	\Err^e&=\mathbb{P}\left[\frac{1}{M_{var}}\mathbf{m}_{i\in\left[M_{inv}, M\right]}^\top Z^e_{var}\le0\mid Y^e=1\right].
\end{align}
Here we assume each element of $Z^e$ and $\mathbf{m}$ are independent with each other, we can have $\Err^e=\mathbb{P}\left[\mathbf{m}_i Z^e_{var, i}\le0\mid Y^e=1\right]$. It is obvious that there is only one situation when the error occurs, i.e., $\mathbf{m}_i=1$ and $Z^e_{var, i}\le0$. Only in this scenario, the sparse training strategy would update the mask to value $0$. Therefore, $\mathbb{P}\left[\mathbf{m}_i=0\right]=1-p^e$. For other cases where $\mathbb{P}\left[Z^e_{var,i}>0\right]=p^e$, the value of each $\mathbf{m}_i$ is randomly initialized and stays intact since there is no error, hence, $\mathbb{P}\left[\mathbf{m}_i=0\right]=\mathbb{P}\left[\mathbf{m}_i=1\right]=\frac{p^e}{2}$. So, for $i\in\left[M_{inv},M\right]$, $\mathbb{P}\left[\mathbf{m}_i=0\right]=1-\frac{p^e}{2}$ and $\mathbb{P}\left[\mathbf{m}_i=1\right]=\frac{p^e}{2}$.

To further bound the error, we denote $\overline{\left[\mathbf{m}Z^e\right]}_{var}=\frac{1}{M_{var}}\mathbf{m}_{i\in\left[M_{inv}, M\right]}^\top Z^e_{var}$, and have:
\begin{align}
	\notag
	\Err^e&=\mathbb{P}\left[\overline{\left[\mathbf{m}Z^e\right]}_{var}\le0\mid Y^e=1\right]\\
	\notag
	&=\mathbb{P}\left[\overline{\left[\mathbf{m}Z^e\right]}_{var}-\mathbb{E}\left[\overline{\left[\mathbf{m}Z^e\right]}_{var}\right]\le-\mathbb{E}\left[\overline{\left[\mathbf{m}Z^e\right]}_{var}\right]\mid Y^e=1\right]\\
	&\le\mathbb{P}\left[\lvert\overline{\left[\mathbf{m}Z^e\right]}_{var}-\mathbb{E}\left[\overline{\left[\mathbf{m}Z^e\right]}_{var}\right]\rvert\ge\mathbb{E}\left[\overline{\left[\mathbf{m}Z^e\right]}_{var}\right]\mid Y^e=1\right].
\end{align}
\begin{align}
	\notag
	\mathbb{E}\left[\overline{\left[\mathbf{m}Z^e\right]}_{var}\right]&=\mathbb{E}\left[\sgn(\frac{1}{M_{var}}\mathbf{m}_{i\in\left[M_{inv}, M\right]}^\top Z^e_{var})\mid Y^e=1\right]\\
	\notag
	&=2\mathbb{P}\left[\frac{1}{M_{var}}\mathbf{m}_{i\in\left[M_{inv}, M\right]}^\top Z^e_{var}>0\mid Y^e=1\right]-1\\
	&=(p^e)^2-1.
\end{align}
Therefore, $\Err^e\le2e^{-2((p^e)^2-1)^2M_{var}}=\mathcal{O}(e^{-(p^e)^4})$. In contrast to the idealized bound that achieves $0$ error in Zhang et al.~\cite{zhang2021can}, when $\theta$ is initialized with the unit norm and given a small $p^e$, the error could be considerably large. This is because the error produced by the variant parameters is largely decided by the probability $p^e$, which could further affect the pruning process. Based on such an intuition, we propose to enhance the pruning strategy by adding an additional regularization that leverages domain knowledge.

Specifically, the regularization considers the errors from distinguishing different distributions using variant parameters:
\begin{align}
	\notag
	&\Err^d=\frac{1}{2}\mathbb{E}_{(X, Y)\sim \mathcal{E}}\left[1-D\hat{D}\right]=\frac{1}{2}\left[1-\mathbb{E}\left[D\hat{D}\right]\right], \\
	&\text{where}\ \hat{D}=\sgn(\theta_{var}^\top Z^e)=\sgn(((1-\mathbf{m})\circ\theta)^\top Z^e).
\end{align}
Similar to Eq.~\ref{eq:appendix_expected_error}, we can have:
\begin{align}
	\notag
	\Err^d&=\mathbb{P}\left[\frac{1}{M}(1-\mathbf{m})^\top Z^e\le0\mid D=1\right]\label{eq:appendix_expected_distribution_error}\\
	&=\mathbb{P}\left[\frac{1}{M_{inv}}(1-\mathbf{m})_{i\in\left[0, M_{inv}\right]}^\top Z^e_{inv}+\frac{1}{M_{var}}(1-\mathbf{m})_{i\in\left[M_{inv}, M\right]}^\top Z^e_{var}\le0\mid D=1\right].
\end{align}
Since all elements in $Z^e_{var}$ equal to $D$, we can have:
\begin{align}
	\Err^d&=\mathbb{P}\left[\frac{1}{M_{inv}}(1-\mathbf{m})_{i\in\left[0, M_{inv}\right]}^\top Z^e_{inv}\le0\mid D=1\right].
\end{align}
Therefore, for $i\in\left[0,M_{inv}\right]$, $\mathbb{P}\left[\mathbf{m}_i=1\right]=1-\frac{q^e}{2}$ and $\mathbb{P}\left[\mathbf{m}_i=0\right]=\frac{q^e}{2}$. As a result, the regularization can complement the mask by finding the invariant parameters with at least probability $1-\frac{q^e}{2}$. Moreover, based on a given sparsity ratio $R=\frac{M_{var}}{M}$, i.e., only $M_{inv}$ elements from $Z^e$ would be selected by $\mathbf{m}$, the erroneous mask that produces classification error can be further constrained from being too much. Particularly, from $\mathbb{P}\left[\mathbf{m}_i=1\right]=1-\frac{q^e}{2},  i\in\left[0,M_{inv}\right]$, we can have $\mathbb{P}\left[\mathbf{m}_i=1\right]=\frac{M_{inv}-(1-\frac{q^e}{2})M_{inv}}{M_{var}}=\frac{q^eM_{inv}}{2M_{var}},  i\in\left[M_{inv}, M\right]$ instead of $\frac{p^e}{2}$ which is calculated based on random initialization. Hence, we can again bound the classification error as:
\begin{align}
	\notag
	\mathbb{E}\left[\overline{\left[\mathbf{m}Z^e\right]}_{var}\right]&=2\mathbb{P}\left[\frac{1}{M_{var}}\mathbf{m}_{i\in\left[M_{inv}, M\right]}^\top Z^e_{var}>0\mid Y^e=1\right]-1\\
	&=p^e\frac{q^eM_{inv}}{M_{var}}-1,
\end{align}
\begin{align}
	\notag
	\Err&^e\le\mathbb{P}\left[\lvert\overline{\left[\mathbf{m}Z^e\right]}_{var}-\mathbb{E}\left[\overline{\left[\mathbf{m}Z^e\right]}_{var}\right]\rvert\ge\mathbb{E}\left[\overline{\left[\mathbf{m}Z^e\right]}_{var}\right]\mid Y^e=1\right]\\
	&\le2e^{-2(\frac{q^eM_{inv}}{M_{var}}p^e-1)^2M_{var}}=\mathcal{O}(e^{-(p^e)^2})
\end{align}

\section{Additional Implementation Details}
\setlength{\intextsep}{0pt}
\setlength{\columnsep}{8pt}
\begin{wrapfigure}{r}{3cm}
	\includegraphics[width=\linewidth]{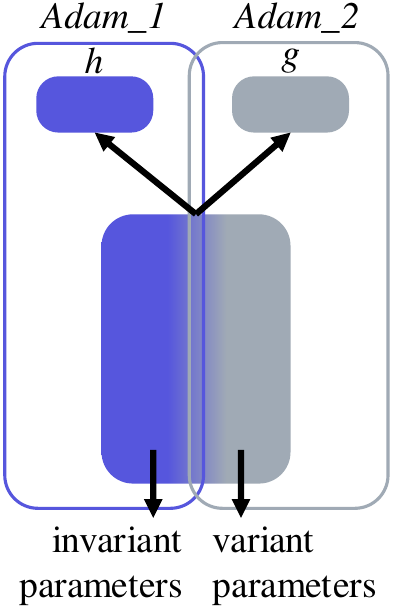}
	\vspace{-6mm}
	\caption{\small Illustration of our optimizers.}
	\label{fig:optim}
\end{wrapfigure}
We have demonstrated the implementation process of our EVIL method, here we provided other details such as the optimization method, hyper-parameters, and the specific experimental setting for each empirical analysis.

To optimize our model, we use Adam~\cite{kingma2014adam} optimizer with an initial learning rate $1e-3$ without weight decay. Moreover, to avoid the conflict between optimizing invariant parameters and variant parameters, we adopt two Adam optimizers, denoted as $Adam_1$ and $Adam_2$, to correspondingly include the invariant parameters and variant parameters. Moreover, $Adam_1$ would include the class prediction head $h$ and $Adam_2$ would include the distribution prediction head $g$, as illustrated in Figure~\ref{fig:optim}. During the training, $Adam_1$ is mainly used to optimize the invariant parameters, but $Adam_2$ is just employed to compute the gradient of variant parameters.

For implementing baseline methods, we mainly follow~\cite{gulrajani2021search} to set the hyper-parameters. Specifically, for DRO, we set $\eta$ as $1e-2$ to update the group importance. For IRM, we set $\lambda=1e2$ to trade off the invariant regularizer. The similar $\lambda$ for penalization from REx is set to $1e1$. For CORAL and MMD, set the trade-off weight $\gamma$ as $1$. For implementing SagNet, we set the weight for adversarial loss as $0.1$. For SAM, we do not use Adaptive SAM~\cite{kwon2021asam} and set the perturbation magnitude $\rho$ as $0.05$.

In the experiments from the main paper, we conducted different experimental settings. Particularly, in Section~\ref{sec:improve_invariant_learning}, we set the sparsity ratio $R = 60\%$, hyper-parameter $\alpha = 0.2$, and $\Delta T = 300$ to implement EVIL. In Section~\ref{sec:comparing_evil_to_sparse}, we keep other hyper-parameters the same but vary the sparsity ratio to evaluate the performance of EVIL under different levels of sparsity. Further, in Section~\ref{sec:performance_analysis}, we keep the same experimental setting as Section~\ref{sec:improve_invariant_learning} except for the parameter sensitivity analysis section, where we carefully tuned the values of $\alpha$ and $\Delta T$ to show their effect on the learning performance.

\section{Additional Experimental Results}
\subsection{Performance on Additional Invariant Learning Methods}
\vspace{3mm}
\begin{wrapfigure}{r}{7cm}
	\small
	\caption{Results on additional invariant learning methods.}
	\setlength{\tabcolsep}{1.5mm}
	\begin{tabular}{lcccc}
		\toprule[1pt]
		Method & MMD & SagNet & Mixstyle & ARM \\ 
		\midrule[0.6pt]
		w/o EVIL & 84.7 & 86.3 & 85.2 & 85.1 \\
		with EVIL & \textbf{85.3} & \textbf{87.1} & \textbf{86.5} & \textbf{86.6} \\ \addlinespace[-0.05cm]
		& ($\pm0.1$) & ($\pm0.2$) & ($\pm0.2$) & ($\pm0.2$) \\
		\toprule[1pt]
	\end{tabular}
	\label{tab:methods}
\end{wrapfigure}
We have discussed several invariant learning methods in the main paper, here we conduct extra experiments on \data{PACS} dataset using additional invariant learning methods to show how EVIL affects their OOD generalization results. Moreover, we conduct experiments using different network architectures to show the effect of EVIL on various learning models.

Concretely, as we have provided results of IRM, REx, DRO, CORAL, and SAM in the main paper, here we implement EVIL using backbone methods including MMD, SagNet, Mixstyle, and ARM. The results on \data{PACS} dataset are shown in Table~\ref{tab:methods}. We can see that our method can still improve the OOD generalization performance which is consistent with the observation in the main paper. Therefore, the proposed EVIL framework is generally effective among various invariant learning methods, which shows great 
deployment practicality of EVIL.

\subsection{Performance on Additional ResNet Architectures}
Moreover, to evaluate the effectiveness of EVIL on different backbone models, we implement the Wide ResNet (WRN)~\cite{zagoruyko2016wide} with varied depths (20, 32, 44, 56, and 110) and train each model from scratch for 500,000 steps to ensure convergence. We also show the result of using ResNet50 pre-trained on ImageNet (Note that due to the pre-training, the performance on ResNet50 would be much better than training from scratch). The comparison between ERM and EVIL is shown in Table~\ref{tab:arch}. Again, we can observe the superiority of EVIL over the baseline method ERM on all investigated architectures. Therefore, we can conclude that the performance improvement brought by EVIL is model-agnostic.

\vspace{3mm}
\begin{table}[h]
	\centering
	\caption{Results on various model architectures. ResNet50 is pre-trained on ImageNet, and other models are trained from scratch.}
	\begin{tabular}{lcccccc}
		\toprule[1pt]
		{\small Arch.} & {\small ResNet50} & {\small WRN-20} & {\small WRN-32} & {\small WRN-44} & {\small  WRN-56} & {\small WRN-110} \\ 
		\midrule[0.6pt]
		ERM & 84.2 & 35.6 & 39.2 & 41.0 & 44.6 & 48.9 \\
		EVIL & \textbf{86.0} & \textbf{37.3} & \textbf{42.5} & \textbf{43.7} & \textbf{47.2} & \textbf{51.4} \\ \addlinespace[-0.1cm]
		& ($\pm0.1$) & ($\pm0.2$) & ($\pm0.3$) & ($\pm0.3$) &($\pm0.2$) & ($\pm0.3$) \\
		\toprule[1pt]
	\end{tabular}
	\label{tab:arch}
\end{table}

\subsection{Optimizing EVIL Using SAM}
\label{sec:ood_sam}
In this section, we first briefly describe the realization of EVIL optimized by SAM for OOD generalization (EVIL-SAM). Then, despite of orthogonality of flatness and OOD generalization as found before~\cite{cha2021swad, rame2022diverse}, we discuss some properties of SAM and demonstrate why combining EVIL and SAM can achieve great performance.

\paragraph{Realization of EVIL-SAM.}
Generally, our EVIL can be optimized using SAM by minimizing the following objectives:
\begin{equation}
	\min_{\theta_{inv}}\max_{\|\bm{\epsilon}\circ\mathbf{m}\|_2\le \rho} \mathcal{L}(\theta_{inv}+\bm{\epsilon}\circ\mathbf{m}; x, y).
	\label{eq:sam_evil}
\end{equation}
Specifically, SAM seeks to compute an optimal parameter perturbation $\bm{\epsilon}^*=\argmax_{\bm{\epsilon}}\mathcal{L}(\theta+\bm{\epsilon}; x, y)$ within $\rho$-radius neighbor that can maximally increase the loss value $\mathcal{L}$. By applying $\bm{\epsilon}^*$, the loss change $\mathcal{L}(\theta+\bm{\epsilon}^*; x, y)-\mathcal{L}(\theta; x, y)$ is denoted as \textit{sharpness} which indicates the flatness of the learned loss function. Intuitively, a flatter loss function often shows better generalization properties, as a slight shift imposed in the input space would not significantly change the loss value. Therefore, SAM has achieved promising in-distribution (ID) generalization performance~\cite{andriushchenko2022towards, kim2022fisher, kwon2021asam, liu2022towards, zhao2022penalizing}. To adopt SAM into EVIL, we just need to apply our mask $\mathbf{m}$ to the parameter perturbation $\bm{\epsilon}$ before computing the optimal $\bm{\epsilon}^*$. This process not only leaves out spurious information but also reduces the computational burden of SAM. As a result, SAM-EVIL can achieve low sharpness for invariant learning.

\begin{table}[t]
	\caption{Comparison of SAM~\cite{foret2020sharpness} and ERM under both ID and OOD situations on DomainBed.}
	\vspace{-3mm}
	\centering
	\begin{tabular}{clccccc|c}
		\toprule[1pt]
		& & {\scriptsize\data{\textbf{PACS}}}  & {\scriptsize\data{\textbf{VLCS}}}  & {\scriptsize\data{\textbf{OfficeHome}}} & {\scriptsize\data{\textbf{TerraInc}}} & {\scriptsize\data{\textbf{DomainNet}}} &  {\scriptsize \textbf{Average}} \\
		\midrule
		\multirow{2}{*}{{\scriptsize ID}} & {\scriptsize ERM}        & 96.6 & 84.7 & 78.9 & 91.3 & 81.4 & 86.5                      \\
		& {\scriptsize SAM} & 97.1 & 86.8 & 82.0 & 93.1 & 85.2 &  88.8                      \\
		\midrule
		\multirow{2}{*}{{\scriptsize OOD}} & {\scriptsize ERM}      & 85.5 & 77.5 & 66.5  & 46.1 & 40.9 &  63.3                      \\
		& {\scriptsize SAM} & 85.8 & 79.4 & 69.6 & 43.3 &  44.3 &  64.5                      \\
		\bottomrule[1pt]
	\end{tabular}
	\label{tab:sam}
\end{table}

\paragraph{Discussion.}
Although SAM has achieved promising ID results, its OOD performance is quite limited~\cite{cha2021swad, rame2022diverse} which is still unexplained. As shown in Table~\ref{tab:sam}, in the ID scenario, SAM shows great effectiveness compared to ERM, but it merely achieves comparable results to ERM in the OOD setting, even worse in some scenarios. In our perspective, the limitation of SAM is caused by erroneously perturbing the variant parameters which encourages fitting to spurious features. Specifically, in OOD problems, the invariant features and spurious ones would activate $\theta_{inv}$ and $\theta_{var}$, respectively. Enforcing robustness (\textit{i.e.}, low sharpness) against perturbation on $\theta_{inv}$ can enhance extracting invariant features. However, by perturbing $\theta_{var}$, low sharpness $\mathcal{L}(\theta+\bm{\epsilon}^*; x, y)-\mathcal{L}(\theta; x, y)$ denotes encouraging the spurious features to bond with the label information. Therefore, SAM cannot extract invariant features as it is sensitive to spurious ones, thus damaging the OOD generalization results. Fortunately, our EVIL can perfectly solve this problem by filtering out the variant parameters which is strongly related to distribution noise. Thus SAM can be further leveraged to enhance the robustness of extracting invariant features. The effectiveness and efficiency of EVIL-SAM are demonstrated in Section~\ref{sec:experiments}.

\begin{table}
	\caption{Comparison of EVIL-SAM with other baseline methods on five datasets from DomainNet.}
	\vspace{-3mm}
	\begin{tabular}{lccccc|c}
		\toprule[1pt]
		{Method} & {\data{\textbf{PACS}}}  & {\data{\textbf{VLCS}}}  & {\data{\textbf{OfficeHome}}} & {\data{\textbf{TerraInc}}} & {\data{\textbf{DomainNet}}} &  { \textbf{Avg.}} \\
		\midrule
		SagNet-SAM  &  86.4  &  78.5  &  69.2  &  49.3  &  40.0  &  64.6 \\
		CORAL-SAM  &  86.6  &  79.0  &  69.3  &  47.9  &  42.1 & 65.0 \\
		MRM-SAM  &  83.9  &  77.1  &  67.0  &  47.4  &  40.6  & 63.2 \\
		SparseIRM-SAM  &  85.2  &  77.4  &  65.6  &  48.5  &  43.1 & 63.9 \\
		EVIL-SAM  &  \textbf{87.8}  &  \textbf{80.1}  & \textbf{ 70.3}  &  \textbf{50.5}  &  \textbf{45.0} & \textbf{66.7}\\
		\bottomrule[1pt]
	\end{tabular}
	\label{tab:justify_sam}
	\vspace{-3mm}
\end{table}

\paragraph{Compare with Other Methods using SAM optimization.}
To further validate our realization that combining EVIL with SAM indeed shows a positive effect, we compare EVIL-SAM to other algorithms as shown in Table~\ref{tab:justify_sam}. We observe that EVIL-SAM achieves the best result among both dense and sparse methods with a significant margin, therefore we can justify our improvement on SAM as more effective than other methods.

\begin{figure}[h]
    \vspace{5mm}
\begin{minipage}[t]{0.5\textwidth}
    \centering
    \includegraphics[width=\textwidth]{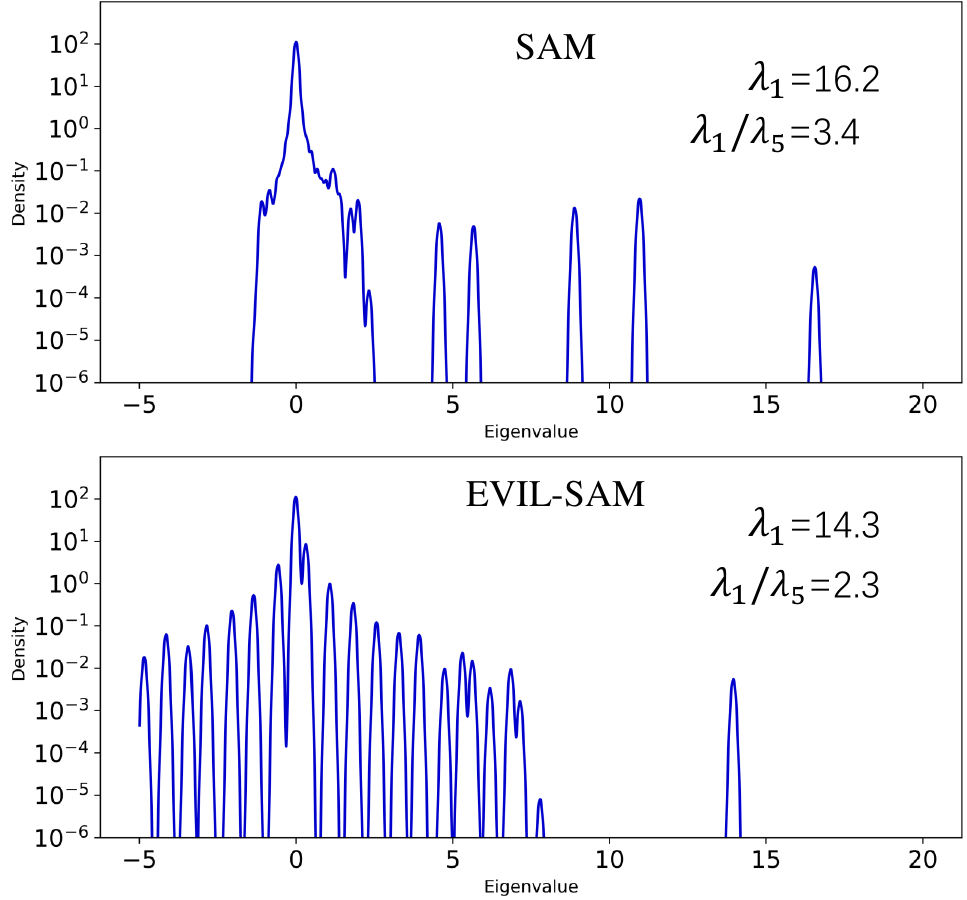}
    \vspace{-5mm}
    \caption{Hessian spectra of SAM and EVIL-SAM.}
    \label{fig:hessian_sam}
\end{minipage}
\begin{minipage}[t]{0.5\textwidth}
    \centering
    \includegraphics[width=\textwidth]{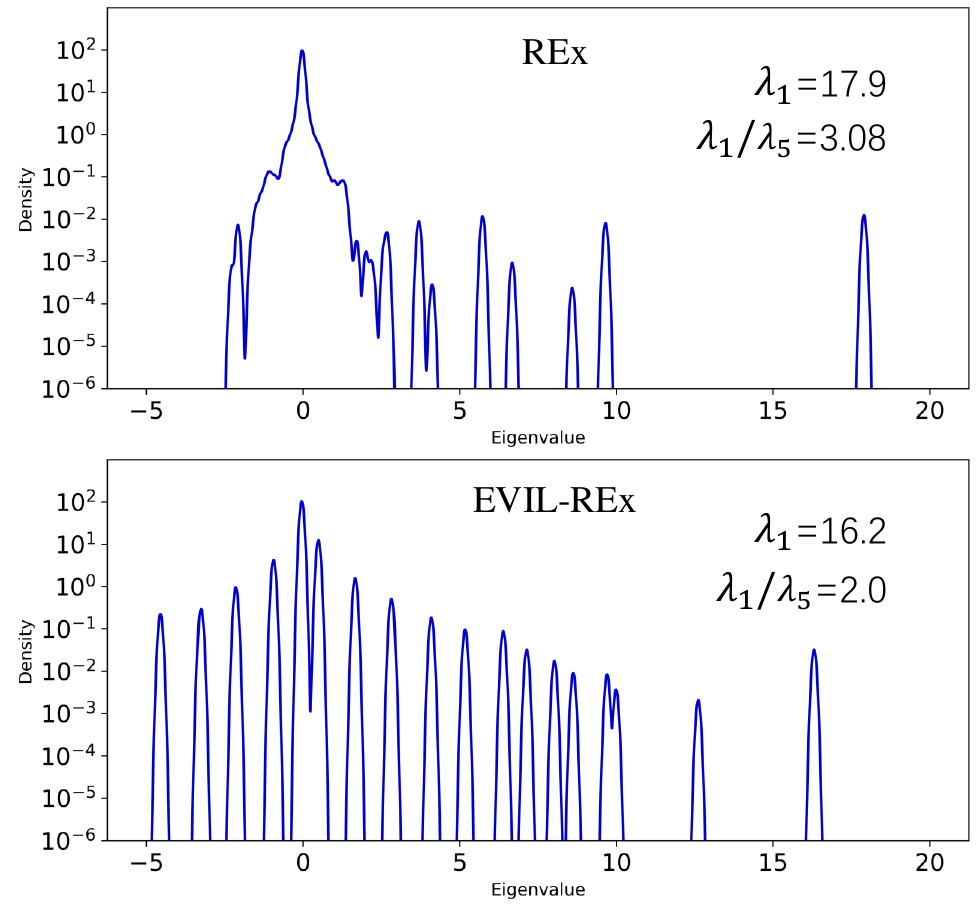}
    \vspace{-5mm}
    \caption{Hessian spectra of REx and EVIL-REx.}
    \label{fig:hessian_rex}
\end{minipage}
\end{figure}

\paragraph{More Hessian Spectra on SAM and REx.}
Since the proposed method shows effective generalization performance, as we have demonstrated in Section~\ref{sec:performance_analysis}, we further validate that the proposed EVIL framework can still help produce improved Hessian spectra when compared to other methods such as SAM and REx. As shown in Figs.~\ref{fig:hessian_sam} and ~\ref{fig:hessian_rex}, We observe the same phenomenon as in the main paper: when combined with EVIL, the largest eigenvalue of both SAM and REx is smaller than its original ones, and the Hessian spectra are more compact when using our EVIL framework. Therefore, we can again conclude that EVIL indeed helps produce flat minima.

\begin{table*}[t!]
	\small
	\centering
		\begin{tabular}{ccccccc}
			\toprule
			& \multicolumn{2}{c}{ImageNet}                                                                                                       & \multicolumn{2}{c}{iWildCam}                                                                                                       & \multicolumn{2}{c}{FMoW}                  \\ \cmidrule(l){2-7} 
			\multirow{-2}{*}{Methods} & ID                  & \multicolumn{1}{c|}{OOD}                                         & ID                  & \multicolumn{1}{c|}{OOD}                 & ID                  & OOD                                          \\ \midrule
			Zeroshot          & 68.3 $\pm$0.0            & \multicolumn{1}{c|}{58.7 $\pm$0.0}           & 8.7 $\pm$0.0             & \multicolumn{1}{c|}{11.0 $\pm$0.0}    & 20.4 $\pm$0.0            & 18.7 $\pm$0.0           \\
			Finetuning       & 82.5 $\pm$0.1          & \multicolumn{1}{c|}{61.3 $\pm$0.1}    & 48.1 $\pm$0.5          & \multicolumn{1}{c|}{35.0 $\pm$0.5}                                  & 68.5 $\pm$0.1          & 39.2 $\pm$0.7          \\
			EVIL     & 81.8 $\pm$0.2          & \multicolumn{1}{c|}{\textbf{62.5} $\pm$0.6}     & 47.6 $\pm$0.8         & \multicolumn{1}{c|}{\textbf{37.4}$\pm$1.2}         & 68.2 $\pm$0.6          & \textbf{41.2} $\pm$1.3          \\ \midrule
		\end{tabular}
	\caption{Performance on ImageNet, iWildCam, and FMoW using CLIP ViT-B/16 as backbone.}
	\label{tab:clip}
	\vspace{-5mm}
\end{table*}

\subsection{Performance on Large-Scale Architecture and Datasets}
In this section, we adopt pretrained CLIP ViT-B/16~\cite{radford2021learning} and conduct finetuning on training datasets from iWildCam, FMoW, and ImageNet, and further test the OOD generalization performance on the split OOD datasets. To extend our sparse training strategy into the CLIP model, we employ a linear layer on top of the ViT backbone and conduct the same pruning strategy by leveraging both class information and domain information. For all datasets, we set the finetuning epoch as 20 and keep the rest of the training parameters the same as described before. The results are shown in Table~\ref{tab:clip}, we can see that although EVIL shows a slight performance drop on ID datasets, which is reasonable since we use fewer parameters than full finetuning, our method achieves the best OOD performance on all three datasets. Specifically, there are $1.2\%$, $2.4\%$, and $2.0\%$ performance gains on ImageNet, iWildCam, and FMoW datasets, respectively. Therefore, the effectiveness and superiority of the proposed EVIL method can be successfully extended to large-scale architectures and datasets.




\newpage
\bibliographystyle{unsrt}
\bibliography{bibliography}

\begin{thebibliography}{10}

\bibitem{rame2022fishr}
Alexandre Rame, Corentin Dancette, and Matthieu Cord.
\newblock Fishr: Invariant gradient variances for out-of-distribution
  generalization.
\newblock In {\em ICML}, pages 18347--18377. PMLR, 2022.

\bibitem{he2016deep}
Kaiming He, Xiangyu Zhang, Shaoqing Ren, and Jian Sun.
\newblock Deep residual learning for image recognition.
\newblock In {\em CVPR}, pages 770--778, 2016.

\bibitem{krizhevsky2009learning}
Alex Krizhevsky, Geoffrey Hinton, et~al.
\newblock Learning multiple layers of features from tiny images.
\newblock 2009.

\bibitem{lecun2015deep}
Yann LeCun, Yoshua Bengio, and Geoffrey Hinton.
\newblock Deep learning.
\newblock {\em nature}, 521(7553):436--444, 2015.

\bibitem{gulrajani2021search}
Ishaan Gulrajani and David Lopez-Paz.
\newblock In search of lost domain generalization.
\newblock In {\em ICLR}, 2021.

\bibitem{li2017deeper}
Da~Li, Yongxin Yang, Yi-Zhe Song, and Timothy~M Hospedales.
\newblock Deeper, broader and artier domain generalization.
\newblock In {\em CVPR}, pages 5542--5550, 2017.

\bibitem{muandet2013domain}
Krikamol Muandet, David Balduzzi, and Bernhard Sch{\"o}lkopf.
\newblock Domain generalization via invariant feature representation.
\newblock In {\em ICML}, pages 10--18. PMLR, 2013.

\bibitem{arjovsky2019invariant}
Martin Arjovsky, L{\'e}on Bottou, Ishaan Gulrajani, and David Lopez-Paz.
\newblock Invariant risk minimization.
\newblock {\em arXiv preprint arXiv:1907.02893}, 2019.

\bibitem{creager2021environment}
Elliot Creager, J{\"o}rn-Henrik Jacobsen, and Richard Zemel.
\newblock Environment inference for invariant learning.
\newblock In {\em ICML}, pages 2189--2200. PMLR, 2021.

\bibitem{lu2021invariant}
Chaochao Lu, Yuhuai Wu, Jos{\'e}~Miguel Hern{\'a}ndez-Lobato, and Bernhard
  Sch{\"o}lkopf.
\newblock Invariant causal representation learning for out-of-distribution
  generalization.
\newblock In {\em ICLR}, 2021.

\bibitem{krueger2021out}
David Krueger, Ethan Caballero, Joern-Henrik Jacobsen, Amy Zhang, Jonathan
  Binas, Dinghuai Zhang, Remi Le~Priol, and Aaron Courville.
\newblock Out-of-distribution generalization via risk extrapolation (rex).
\newblock In {\em ICML}, pages 5815--5826. PMLR, 2021.

\bibitem{frankle2019lottery}
Jonathan Frankle and Michael Carbin.
\newblock The lottery ticket hypothesis: Finding sparse, trainable neural
  networks.
\newblock In {\em ICLR}, 2019.

\bibitem{frankle2020linear}
Jonathan Frankle, Gintare~Karolina Dziugaite, Daniel Roy, and Michael Carbin.
\newblock Linear mode connectivity and the lottery ticket hypothesis.
\newblock In {\em ICML}, pages 3259--3269. PMLR, 2020.

\bibitem{malach2020proving}
Eran Malach, Gilad Yehudai, Shai Shalev-Schwartz, and Ohad Shamir.
\newblock Proving the lottery ticket hypothesis: Pruning is all you need.
\newblock In {\em ICML}, pages 6682--6691. PMLR, 2020.

\bibitem{morcos2019one}
Ari Morcos, Haonan Yu, Michela Paganini, and Yuandong Tian.
\newblock One ticket to win them all: generalizing lottery ticket
  initializations across datasets and optimizers.
\newblock In {\em NeurIPS}, volume~32, 2019.

\bibitem{zhang2021can}
Dinghuai Zhang, Kartik Ahuja, Yilun Xu, Yisen Wang, and Aaron Courville.
\newblock Can subnetwork structure be the key to out-of-distribution
  generalization?
\newblock In {\em ICML}, pages 12356--12367. PMLR, 2021.

\bibitem{zhou2022sparse}
Xiao Zhou, Yong Lin, Weizhong Zhang, and Tong Zhang.
\newblock Sparse invariant risk minimization.
\newblock In {\em ICML}, pages 27222--27244. PMLR, 2022.

\bibitem{mitrovic2020representation}
Jovana Mitrovic, Brian McWilliams, Jacob Walker, Lars Buesing, and Charles
  Blundell.
\newblock Representation learning via invariant causal mechanisms.
\newblock {\em arXiv preprint arXiv:2010.07922}, 2020.

\bibitem{von2021self}
Julius Von~K{\"u}gelgen, Yash Sharma, Luigi Gresele, Wieland Brendel, Bernhard
  Sch{\"o}lkopf, Michel Besserve, and Francesco Locatello.
\newblock Self-supervised learning with data augmentations provably isolates
  content from style.
\newblock In {\em NeurIPS}, volume~34, 2021.

\bibitem{zhang2013domain}
Kun Zhang, Bernhard Sch{\"o}lkopf, Krikamol Muandet, and Zhikun Wang.
\newblock Domain adaptation under target and conditional shift.
\newblock In {\em ICML}, pages 819--827. PMLR, 2013.

\bibitem{huang2020causal}
Biwei Huang, Kun Zhang, Jiji Zhang, Joseph Ramsey, Ruben Sanchez-Romero, Clark
  Glymour, and Bernhard Sch{\"o}lkopf.
\newblock Causal discovery from heterogeneous/nonstationary data.
\newblock {\em The Journal of Machine Learning Research}, 21(1):3482--3534,
  2020.

\bibitem{suter2019robustly}
Raphael Suter, Djordje Miladinovic, Bernhard Sch{\"o}lkopf, and Stefan Bauer.
\newblock Robustly disentangled causal mechanisms: Validating deep
  representations for interventional robustness.
\newblock In {\em ICML}, pages 6056--6065. PMLR, 2019.

\bibitem{huang2021universal}
Zhuo Huang, Chao Xue, Bo~Han, Jian Yang, and Chen Gong.
\newblock Universal semi-supervised learning.
\newblock In {\em NeurIPS}, volume~34, 2021.

\bibitem{huang2022they}
Zhuo Huang, Jian Yang, and Chen Gong.
\newblock They are not completely useless: Towards recycling transferable
  unlabeled data for class-mismatched semi-supervised learning.
\newblock {\em IEEE Transactions on Multimedia}, 2022.

\bibitem{bai2022rsa}
Yingbin Bai, Erkun Yang, Zhaoqing Wang, Yuxuan Du, Bo~Han, Cheng Deng, Dadong
  Wang, and Tongliang Liu.
\newblock Rsa: Reducing semantic shift from aggressive augmentations for
  self-supervised learning.
\newblock In {\em NeurIPS}, volume~35, pages 21128--21141, 2022.

\bibitem{hu2018does}
Weihua Hu, Gang Niu, Issei Sato, and Masashi Sugiyama.
\newblock Does distributionally robust supervised learning give robust
  classifiers?
\newblock In {\em ICML}, pages 2029--2037. PMLR, 2018.

\bibitem{huang2023robust}
Zhuo Huang, Miaoxi Zhu, Xiaobo Xia, Li~Shen, Jun Yu, Chen Gong, Bo~Han, Bo~Du,
  and Tongliang Liu.
\newblock Robust generalization against photon-limited corruptions via
  worst-case sharpness minimization.
\newblock In {\em Proceedings of the IEEE/CVF Conference on Computer Vision and
  Pattern Recognition}, pages 16175--16185, 2023.

\bibitem{sagawa2019distributionally}
Shiori Sagawa, Pang~Wei Koh, Tatsunori~B Hashimoto, and Percy Liang.
\newblock Distributionally robust neural networks for group shifts: On the
  importance of regularization for worst-case generalization.
\newblock In {\em International Conference on Learning Representations}, 2020.

\bibitem{wang2023doe}
Qizhou Wang, Junjie Ye, Feng Liu, Quanyu Dai, Marcus Kalander, Tongliang Liu,
  Jianye Hao, and Bo~Han.
\newblock Out-of-distribution detection with implicit outlier transformation.
\newblock In {\em ICLR}, 2023.

\bibitem{foret2020sharpness}
Pierre Foret, Ariel Kleiner, Hossein Mobahi, and Behnam Neyshabur.
\newblock Sharpness-aware minimization for efficiently improving
  generalization.
\newblock In {\em ICLR}, 2020.

\bibitem{cha2021swad}
Junbum Cha, Sanghyuk Chun, Kyungjae Lee, Han-Cheol Cho, Seunghyun Park, Yunsung
  Lee, and Sungrae Park.
\newblock Swad: Domain generalization by seeking flat minima.
\newblock In {\em NeurIPS}, volume~34, pages 22405--22418, 2021.

\bibitem{huang2023FlatMatch}
Zhuo Huang, Li~Shen, Jun Yu, Bo~Han, and Tongliang Liu.
\newblock Flatmatch: Bridging labeled data and unlabeled data with
  cross-sharpness for semi-supervised learning.
\newblock In {\em Advances in Neural Information Processing Systems}, 2023.

\bibitem{kang2023unleashing}
Hui Kang, Sheng Liu, Huaxi Huang, Jun Yu, Bo~Han, Dadong Wang, and Tongliang
  Liu.
\newblock Unleashing the potential of regularization strategies in learning
  with noisy labels.
\newblock {\em arXiv preprint arXiv:2307.05025}, 2023.

\bibitem{xia2023combating}
Xiaobo Xia, Bo~Han, Yibing Zhan, Jun Yu, Mingming Gong, Chen Gong, and
  Tongliang Liu.
\newblock Combating noisy labels with sample selection by mining
  high-discrepancy examples.
\newblock In {\em ICCV}, pages 1833--1843, 2023.

\bibitem{xia2021sample}
Xiaobo Xia, Tongliang Liu, Bo~Han, Mingming Gong, Jun Yu, Gang Niu, and Masashi
  Sugiyama.
\newblock Sample selection with uncertainty of losses for learning with noisy
  labels.
\newblock {\em arXiv preprint arXiv:2106.00445}, 2021.

\bibitem{glymour2016causal}
Madelyn Glymour, Judea Pearl, and Nicholas~P Jewell.
\newblock {\em Causal inference in statistics: A primer}.
\newblock 2016.

\bibitem{pearl2009causality}
Judea Pearl.
\newblock {\em Causality}.
\newblock 2009.

\bibitem{huang2023harnessing}
Zhuo Huang, Xiaobo Xia, Li~Shen, Bo~Han, Mingming Gong, Chen Gong, and
  Tongliang Liu.
\newblock Harnessing out-of-distribution examples via augmenting content and
  style.
\newblock In {\em ICLR}, 2023.

\bibitem{wang2022exploring}
Zhaoqing Wang, Qiang Li, Guoxin Zhang, Pengfei Wan, Wen Zheng, Nannan Wang,
  Mingming Gong, and Tongliang Liu.
\newblock Exploring set similarity for dense self-supervised representation
  learning.
\newblock In {\em CVPR}, pages 16590--16599, 2022.

\bibitem{gong2016domain}
Mingming Gong, Kun Zhang, Tongliang Liu, Dacheng Tao, Clark Glymour, and
  Bernhard Sch{\"o}lkopf.
\newblock Domain adaptation with conditional transferable components.
\newblock In {\em ICML}, pages 2839--2848. PMLR, 2016.

\bibitem{wang2023learning}
Qizhou Wang, Zhen Fang, Yonggang Zhang, Feng Liu, Yixuan Li, and Bo~Han.
\newblock Learning to augment distributions for out-of-distribution detection.
\newblock In {\em NeurIPS}, 2023.

\bibitem{wang2022watermark}
Qizhou Wang, Feng Liu, Yonggang Zhang, Jing Zhang, Chen Gong, Tongliang Liu,
  and Bo~Han.
\newblock Watermarking for out-of-distribution detection.
\newblock In {\em NeurIPS}, 2022.

\bibitem{liu2021learning}
Chang Liu, Xinwei Sun, Jindong Wang, Haoyue Tang, Tao Li, Tao Qin, Wei Chen,
  and Tie-Yan Liu.
\newblock Learning causal semantic representation for out-of-distribution
  prediction.
\newblock In {\em NeurIPS}, volume~34, 2021.

\bibitem{sun2021recovering}
Xinwei Sun, Botong Wu, Xiangyu Zheng, Chang Liu, Wei Chen, Tao Qin, and Tie-Yan
  Liu.
\newblock Recovering latent causal factor for generalization to distributional
  shifts.
\newblock In {\em NeurIPS}, volume~34, 2021.

\bibitem{tsipras2018robustness}
Dimitris Tsipras, Shibani Santurkar, Logan Engstrom, Alexander Turner, and
  Aleksander Madry.
\newblock Robustness may be at odds with accuracy.
\newblock In {\em ICLR}, 2018.

\bibitem{rosenfeld2020risks}
Elan Rosenfeld, Pradeep Ravikumar, and Andrej Risteski.
\newblock The risks of invariant risk minimization.
\newblock {\em arXiv preprint arXiv:2010.05761}, 2020.

\bibitem{lee2018snip}
Namhoon Lee, Thalaiyasingam Ajanthan, and Philip~HS Torr.
\newblock Snip: Single-shot network pruning based on connection sensitivity.
\newblock In {\em ICLR}, 2019.

\bibitem{evci2020rigging}
Utku Evci, Trevor Gale, Jacob Menick, Pablo~Samuel Castro, and Erich Elsen.
\newblock Rigging the lottery: Making all tickets winners.
\newblock In {\em ICML}, pages 2943--2952. PMLR, 2020.

\bibitem{sung2021training}
Yi-Lin Sung, Varun Nair, and Colin~A Raffel.
\newblock Training neural networks with fixed sparse masks.
\newblock In {\em NeurIPS}, volume~34, pages 24193--24205, 2021.

\bibitem{chen2021lottery}
Tianlong Chen, Jonathan Frankle, Shiyu Chang, Sijia Liu, Yang Zhang, Michael
  Carbin, and Zhangyang Wang.
\newblock The lottery tickets hypothesis for supervised and self-supervised
  pre-training in computer vision models.
\newblock In {\em CVPR}, pages 16306--16316, 2021.

\bibitem{dettmers2019sparse}
Tim Dettmers and Luke Zettlemoyer.
\newblock Sparse networks from scratch: Faster training without losing
  performance.
\newblock {\em arXiv preprint arXiv:1907.04840}, 2019.

\bibitem{csordas2020neural}
R{\'o}bert Csord{\'a}s, Sjoerd van Steenkiste, and J{\"u}rgen Schmidhuber.
\newblock Are neural nets modular? inspecting functional modularity through
  differentiable weight masks.
\newblock In {\em ICLR}, 2020.

\bibitem{louizoslearning}
Christos Louizos, Max Welling, and Diederik~P Kingma.
\newblock Learning sparse neural networks through l\_0 regularization.
\newblock In {\em ICLR}, 2018.

\bibitem{liu2022unreasonable}
Shiwei Liu, Tianlong Chen, Xiaohan Chen, Li~Shen, Decebal~Constantin Mocanu,
  Zhangyang Wang, and Mykola Pechenizkiy.
\newblock The unreasonable effectiveness of random pruning: Return of the most
  naive baseline for sparse training.
\newblock In {\em ICLR}, 2022.

\bibitem{sun2016deep}
Baochen Sun and Kate Saenko.
\newblock Deep coral: Correlation alignment for deep domain adaptation.
\newblock In {\em ECCV}, pages 443--450. Springer, 2016.

\bibitem{cha2022domain}
Junbum Cha, Kyungjae Lee, Sungrae Park, and Sanghyuk Chun.
\newblock Domain generalization by mutual-information regularization with
  pre-trained models.
\newblock In {\em ECCV}, pages 440--457. Springer, 2022.

\bibitem{ghifary2015domain}
Muhammad Ghifary, W~Bastiaan Kleijn, Mengjie Zhang, and David Balduzzi.
\newblock Domain generalization for object recognition with multi-task
  autoencoders.
\newblock In {\em ICCV}, pages 2551--2559, 2015.

\bibitem{fang2013unbiased}
Chen Fang, Ye~Xu, and Daniel~N Rockmore.
\newblock Unbiased metric learning: On the utilization of multiple datasets and
  web images for softening bias.
\newblock In {\em CVPR}, pages 1657--1664, 2013.

\bibitem{venkateswara2017deep}
Hemanth Venkateswara, Jose Eusebio, Shayok Chakraborty, and Sethuraman
  Panchanathan.
\newblock Deep hashing network for unsupervised domain adaptation.
\newblock In {\em CVPR}, pages 5018--5027, 2017.

\bibitem{beery2018recognition}
Sara Beery, Grant Van~Horn, and Pietro Perona.
\newblock Recognition in terra incognita.
\newblock In {\em ECCV}, pages 456--473, 2018.

\bibitem{peng2019moment}
Xingchao Peng, Qinxun Bai, Xide Xia, Zijun Huang, Kate Saenko, and Bo~Wang.
\newblock Moment matching for multi-source domain adaptation.
\newblock In {\em ICCV}, pages 1406--1415, 2019.

\bibitem{koh2021wilds}
Pang~Wei Koh, Shiori Sagawa, Henrik Marklund, Sang~Michael Xie, Marvin Zhang,
  Akshay Balsubramani, Weihua Hu, Michihiro Yasunaga, Richard~Lanas Phillips,
  Irena Gao, et~al.
\newblock Wilds: A benchmark of in-the-wild distribution shifts.
\newblock In {\em ICML}, pages 5637--5664. PMLR, 2021.

\bibitem{russakovsky2015imagenet}
Olga Russakovsky, Jia Deng, Hao Su, Jonathan Krause, Sanjeev Satheesh, Sean Ma,
  Zhiheng Huang, Andrej Karpathy, Aditya Khosla, Michael Bernstein, et~al.
\newblock Imagenet large scale visual recognition challenge.
\newblock {\em International journal of computer vision}, 115(3):211--252,
  2015.

\bibitem{recht2019imagenet}
Benjamin Recht, Rebecca Roelofs, Ludwig Schmidt, and Vaishaal Shankar.
\newblock Do imagenet classifiers generalize to imagenet?
\newblock In {\em ICML}, pages 5389--5400. PMLR, 2019.

\bibitem{hendrycks2021many}
Dan Hendrycks, Steven Basart, Norman Mu, Saurav Kadavath, Frank Wang, Evan
  Dorundo, Rahul Desai, Tyler Zhu, Samyak Parajuli, Mike Guo, et~al.
\newblock The many faces of robustness: A critical analysis of
  out-of-distribution generalization.
\newblock In {\em ICCV}, pages 8340--8349, 2021.

\bibitem{hendrycks2021natural}
Dan Hendrycks, Kevin Zhao, Steven Basart, Jacob Steinhardt, and Dawn Song.
\newblock Natural adversarial examples.
\newblock In {\em CVPR}, pages 15262--15271, 2021.

\bibitem{wang2019learning}
Haohan Wang, Songwei Ge, Zachary Lipton, and Eric~P Xing.
\newblock Learning robust global representations by penalizing local predictive
  power.
\newblock In {\em NeurIPS}, volume~32, 2019.

\bibitem{barbu2019objectnet}
Andrei Barbu, David Mayo, Julian Alverio, William Luo, Christopher Wang, Dan
  Gutfreund, Josh Tenenbaum, and Boris Katz.
\newblock Objectnet: A large-scale bias-controlled dataset for pushing the
  limits of object recognition models.
\newblock In {\em NeurIPS}, volume~32, 2019.

\bibitem{radford2021learning}
Alec Radford, Jong~Wook Kim, Chris Hallacy, Aditya Ramesh, Gabriel Goh,
  Sandhini Agarwal, Girish Sastry, Amanda Askell, Pamela Mishkin, Jack Clark,
  Gretchen Krueger, and Ilya Sutskever.
\newblock Learning transferable visual models from natural language
  supervision.
\newblock In {\em ICML}, pages 8748--8763, 2021.

\bibitem{rame2022diverse}
Alexandre Rame, Matthieu Kirchmeyer, Thibaud Rahier, Alain Rakotomamonjy,
  Patrick Gallinari, and Matthieu Cord.
\newblock Diverse weight averaging for out-of-distribution generalization.
\newblock In {\em NeurIPS}, 2022.

\bibitem{izmailov2018averaging}
Pavel Izmailov, Dmitrii Podoprikhin, Timur Garipov, Dmitry Vetrov, and
  Andrew~Gordon Wilson.
\newblock Averaging weights leads to wider optima and better generalization.
\newblock {\em arXiv preprint arXiv:1803.05407}, 2018.

\bibitem{bai2021me}
Yingbin Bai and Tongliang Liu.
\newblock Me-momentum: Extracting hard confident examples from noisily labeled
  data.
\newblock In {\em ICCV}, pages 9312--9321, 2021.

\bibitem{xu2020adversarial}
Minghao Xu, Jian Zhang, Bingbing Ni, Teng Li, Chengjie Wang, Qi~Tian, and
  Wenjun Zhang.
\newblock Adversarial domain adaptation with domain mixup.
\newblock In {\em AAAI}, volume~34, pages 6502--6509, 2020.

\bibitem{li2018learning}
Da~Li, Yongxin Yang, Yi-Zhe Song, and Timothy Hospedales.
\newblock Learning to generalize: Meta-learning for domain generalization.
\newblock In {\em AAI}, volume~32, 2018.

\bibitem{li2018domain}
Haoliang Li, Sinno~Jialin Pan, Shiqi Wang, and Alex~C Kot.
\newblock Domain generalization with adversarial feature learning.
\newblock In {\em CVPR}, pages 5400--5409, 2018.

\bibitem{ganin2015unsupervised}
Yaroslav Ganin and Victor Lempitsky.
\newblock Unsupervised domain adaptation by backpropagation.
\newblock In {\em ICML}, pages 1180--1189. PMLR, 2015.

\bibitem{long2018conditional}
Mingsheng Long, Zhangjie Cao, Jianmin Wang, and Michael~I Jordan.
\newblock Conditional adversarial domain adaptation.
\newblock In {\em NeurIPS}, 2018.

\bibitem{blanchard2021domain}
Gilles Blanchard, Aniket~Anand Deshmukh, {\"U}run Dogan, Gyemin Lee, and
  Clayton Scott.
\newblock Domain generalization by marginal transfer learning.
\newblock {\em The Journal of Machine Learning Research}, 22(1):46--100, 2021.

\bibitem{nam2021reducing}
Hyeonseob Nam, HyunJae Lee, Jongchan Park, Wonjun Yoon, and Donggeun Yoo.
\newblock Reducing domain gap by reducing style bias.
\newblock In {\em CVPR}, pages 8690--8699, 2021.

\bibitem{zhang2020adaptive}
Marvin Zhang, Henrik Marklund, Abhishek Gupta, Sergey Levine, and Chelsea Finn.
\newblock Adaptive risk minimization: A meta-learning approach for tackling
  group shift.
\newblock {\em arXiv preprint arXiv:2007.02931}, 8:9, 2020.

\bibitem{huang2020self}
Zeyi Huang, Haohan Wang, Eric~P Xing, and Dong Huang.
\newblock Self-challenging improves cross-domain generalization.
\newblock In {\em ECCV}, pages 124--140. Springer, 2020.

\bibitem{zhou2021domain}
Kaiyang Zhou, Yongxin Yang, Yu~Qiao, and Tao Xiang.
\newblock Domain generalization with mixstyle.
\newblock In {\em ICLR}, 2021.

\bibitem{vapnik1999overview}
Vladimir~N Vapnik.
\newblock An overview of statistical learning theory.
\newblock {\em IEEE transactions on neural networks}, 10(5):988--999, 1999.

\bibitem{ghorbani2019investigation}
Behrooz Ghorbani, Shankar Krishnan, and Ying Xiao.
\newblock An investigation into neural net optimization via hessian eigenvalue
  density.
\newblock In {\em ICML}, pages 2232--2241. PMLR, 2019.

\bibitem{fort2021exploring}
Stanislav Fort, Jie Ren, and Balaji Lakshminarayanan.
\newblock Exploring the limits of out-of-distribution detection.
\newblock In {\em NeurIPS}, volume~34, 2021.

\bibitem{kingma2014adam}
Diederik~P Kingma and Jimmy Ba.
\newblock Adam: A method for stochastic optimization.
\newblock {\em arXiv preprint arXiv:1412.6980}, 2014.

\bibitem{kwon2021asam}
Jungmin Kwon, Jeongseop Kim, Hyunseo Park, and In~Kwon Choi.
\newblock Asam: Adaptive sharpness-aware minimization for scale-invariant
  learning of deep neural networks.
\newblock In {\em ICML}, pages 5905--5914. PMLR, 2021.

\bibitem{zagoruyko2016wide}
Sergey Zagoruyko and Nikos Komodakis.
\newblock Wide residual networks.
\newblock In {\em BMVC}, 2016.

\bibitem{andriushchenko2022towards}
Maksym Andriushchenko and Nicolas Flammarion.
\newblock Towards understanding sharpness-aware minimization.
\newblock In {\em ICML}, pages 639--668. PMLR, 2022.

\bibitem{kim2022fisher}
Minyoung Kim, Da~Li, Shell~X Hu, and Timothy Hospedales.
\newblock Fisher sam: Information geometry and sharpness aware minimisation.
\newblock In {\em ICML}, pages 11148--11161. PMLR, 2022.

\bibitem{liu2022towards}
Yong Liu, Siqi Mai, Xiangning Chen, Cho-Jui Hsieh, and Yang You.
\newblock Towards efficient and scalable sharpness-aware minimization.
\newblock In {\em CVPR}, pages 12360--12370, 2022.

\bibitem{zhao2022penalizing}
Yang Zhao, Hao Zhang, and Xiuyuan Hu.
\newblock Penalizing gradient norm for efficiently improving generalization in
  deep learning.
\newblock In {\em ICML}, 2022.

\end{thebibliography}

\end{document}